\documentclass[]{fairmeta}
\usepackage{graphicx}
\usepackage{amsmath}
\usepackage{amssymb}
\usepackage{booktabs}
\usepackage{adjustbox}
\usepackage{multirow}
\usepackage{arydshln}
\usepackage{colortbl}
\usepackage{overpic}
\usepackage{amsthm}

\usepackage{amsfonts}       % blackboard math symbols
\usepackage{nicefrac}       % compact symbols for 1/2, etc.
\usepackage{microtype}      % microtypography

\usepackage{scalerel}

\usepackage{makecell}

\usepackage{algorithm}
\usepackage{algorithmic}
\usepackage{enumitem}

\usepackage{nicefrac}
\usepackage{multirow}
\usepackage{booktabs}
\usepackage{float}
\usepackage{wrapfig}
\usepackage{bbding}
\usepackage{pifont}
\usepackage{color, colortbl}

\definecolor{gray}{gray}{0.9}
\newcommand{\ssymbol}[1]{^{\@fnsymbol{#1}}}
\def\R{\mathbb{R}}
\newcommand{\Lleft}{\left(}
\newcommand{\Rright}{\right)}

\newtheorem{lemma}{Lemma}

\title{Efficient Track Anything}

\author[1,\dagger]{Yunyang Xiong}
\author[1,2]{Chong Zhou}
\author[1]{Xiaoyu Xiang}
\author[1]{Lemeng Wu}
\author[1]{Chenchen Zhu}
\author[1]{Zechun Liu}
\author[1]{Saksham Suri}
\author[1]{Balakrishnan Varadarajan}
\author[1]{Ramya Akula}
\author[1]{Forrest Iandola}
\author[1,\dagger]{Raghuraman Krishnamoorthi}
\author[1,\dagger]{Bilge Soran}
\author[1,\dagger]{Vikas Chandra}

\affiliation[1]{Meta AI}
\affiliation[2]{Nanyang Technological University}

\contribution[\dagger]{Project lead}

\abstract{Segment Anything Model 2 (SAM 2) has emerged as a powerful tool for video object segmentation and tracking anything. Key components of SAM 2 that drive the impressive video object segmentation performance include a large multistage image encoder for frame feature extraction and a memory mechanism that stores memory contexts from past frames to help current frame segmentation. The high computation complexity of multistage image encoder and memory module has limited its applications in real-world tasks, e.g., video object segmentation on mobile devices. To address this limitation, we propose EfficientTAMs, lightweight track anything models that produce high-quality results with low latency and model size. Our idea is based on revisiting the plain, nonhierarchical Vision Transformer (ViT) as an image encoder for video object segmentation, and introducing an efficient memory module, which reduces the complexity for both frame feature extraction and memory computation for current frame segmentation. We take vanilla lightweight ViTs and efficient memory module to build EfficientTAMs, and train the models on SA-1B and SA-V datasets for video object segmentation and track anything tasks. We evaluate on multiple video segmentation benchmarks including semi-supervised VOS and promptable video segmentation, and find that our proposed EfficientTAM with vanilla ViT perform comparably to SAM 2 model (HieraB+SAM 2) with $\sim$2x speedup on A100 and $\sim$2.4x  parameter reduction. On segment anything image tasks, our EfficientTAMs also perform favorably over original SAM with $\sim$20x  speedup on A100 and $\sim$20x  parameter reduction. On mobile devices such as iPhone 15 Pro Max, our EfficientTAMs can run at $\sim$10 FPS for performing video object segmentation with reasonable quality, highlighting the capability of small models for on-device video object segmentation applications.}

\correspondence{\email{yunyang@meta.com}}

\metadata[Project]{\url{https://yformer.github.io/efficient-track-anything/}}

\begin{document}

\maketitle

% \vspace{-2mm}
\section{Introduction}
\label{sec:intro}
% \vspace{-3mm}
\begin{figure}[t]
    \centering
    % \vspace{-3mm}
    \includegraphics[width=\linewidth]{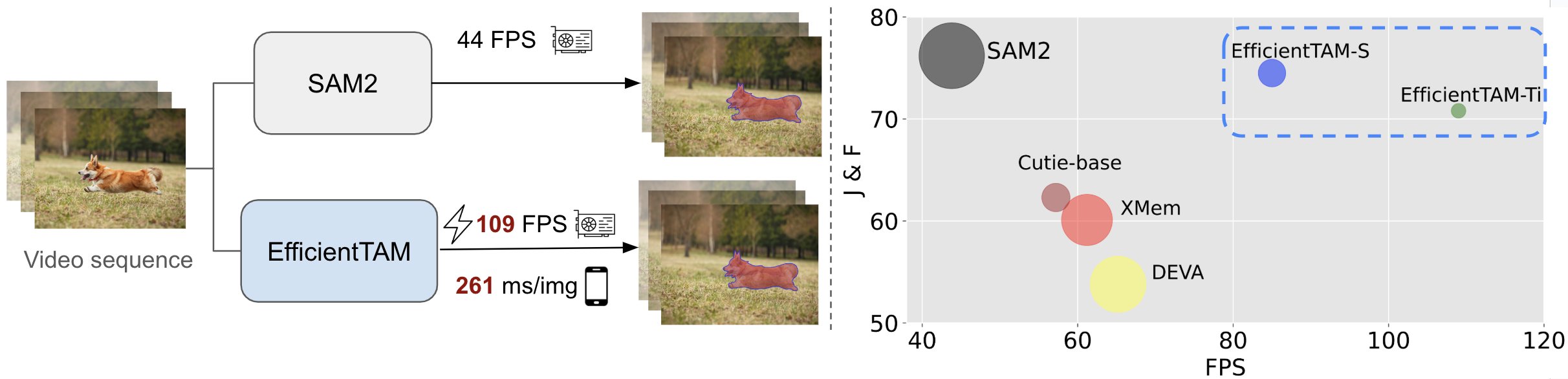}
    \caption{Comparative analysis. (Left) Speed comparison between EfficientTAM and SAM 2 on a single NVIDIA A100 GPU. While SAM 2 is challenging for on-device deployment, our EfficientTAM can run 261 ms per frame on iPhone 15 Pro Max. (Right) 
    FPS/Parameter/Performance comparison of EfficientTAM, SAM 2, and other efficient models for zero-shot video object segmentation on SA-V test. We benchmark FPS (frames per second) of all models with 1024 × 1024 input resolution on a single NVIDIA A100.}
    \label{fig:throughput}
% \vspace{-3mm}
\end{figure}

Segment Anything Model 2 (SAM 2)~\citep{ravi2024sam} is a foundational model for unified image and video object segmentation, achieving state-of-the-art performance in various segmentation tasks such as zero-shot image segmentation~\citep{kirillov2023segment,chen2023semantic,deng2023segment,chen2023sam}, semi-supervised video object segmentation~\citep{pont20172017,xu2018youtube,oh2019video,bhat2020learning,robinson2020learning,li2022recurrent,yang2022decoupling,cheng2022xmem,zhang2023joint,wang2023look,wu2023scalable,cheng2024putting,yang2024scalable}, interactive video segmentation~\citep{caelles20182018,heo2020interactive,cheng2021modular,homayounfar2021videoclick,yang2023track,cheng2023segment,rajivc2023segment,cheng2024putting,delatolas2024learning}, and other real-world applications~\citep{zhang2024evf,xiong2024sam2,shen2024performance,zhang2024sam2,ding2024sam2long,qiu2024ded,tang2024segment,zhou2024sam2}. SAM 2 uses a multistage image encoder  to extract hierarchical frame features and introduces a memory module to cross-attend to both current frame features and stored memories from observed frames for consistent object segmentation across frames and interactive tracking in videos. 
%This makes SAM 2 a vision foundation model and enables its applications beyond vision. 

Despite these advantages, SAM 2 is not efficient for mobile deployment, particularly because the large image encoder (e.g., HieraB+) and memory module are expensive. The default image encoder of SAM 2, HieraB+~\citep{ryali2023hiera}, is parameter inefficient, e.g., $\sim$80M parameters. While SAM 2 provides a tiny version, it has a running time of 43.8 FPS comparable to 47.2 FPS of the default SAM 2 model, due to the hierarchical image encoder. Additionally, that the memory tokens (e.g., a concatenation of spatial memory tokens and object pointer tokens) are long, e.g., $\sim$30K, which hurts the efficiency of the memory module with cross-attention. 
%Consequently, SAM 2 has high memory and computational costs when performing video object segmentation and track anything tasks, which makes it challenging for on-device applications. 

In this paper, we revisit plain, nonhierarchical image encoder for video object segmentation and tracking anything. We propose using a lightweight vanilla ViT image encoder (e.g., ViT-Tiny/-Small\citep{touvron2021training}) as EfficientSAMs\citep{xiong2024efficientsam} did to reduce the complexity of SAM 2 while maintaining decent performance. Further, we propose an efficient cross-attention method for accelerating the memory module. This is achieved by leveraging the underlying structure of memory spatial tokens. We observed that the memory spatial tokens have strong locality and a coarser representation of memory spatial tokens can be a good proxy for performing cross-attention. We show that this yields a good alternative to the original memory module. 

To evaluate our method, we conduct extensive experiments across video and image segmentation benchmarks, including MOSE, DAVIS, LVOS, and SA-V for video segmentation, and SA-23 for image segmentation. Our EfficientTAM outperforms strong semi-supervised video object segmentation methods such as Cutie-base, XMem, and DEVA while being more efficient. Compared with SAM 2, our EfficientTAM is comparable, e.g., 74.5\% vs 74.7\% on SA-V test dataset, with $\sim$ 2x reduced FPS. On image segmentation benchmark, SA-23, our EfficientTAM achieves 60.7\% accuracy for zero-shot image segmentation compared to 59.1\% accuracy for SAM and and 61.9\% for SAM 2. We also benchmarked our EfficientTAM model on iPhone 15 Pro Max, which can run $\sim$ 10 frames per second with reasonable video segmentation performance.
% compared to x frames per second for SAM and SAM 2. 

Our main contributions can be summarized as follows:
\begin{itemize}[noitemsep,topsep=0pt,leftmargin=0.5cm]
    \item 
    We revisit using plain, non-hierarchical image encoder, ViT-Tiny/-Small for video object segmentation and show that vanilla ViT can achieve competing performance comparing to SAM 2 with hierarchical image encoder.

    \item We propose an efficient memory cross-attention by exploiting the underlying memory spatial token structure and demonstrate the favorable performance.

    \item We deliver EfficientTAMs, lightweight video object segmentation and track anything models with state-of-the-art quality-efficiency tradeoffs (\cref{fig:throughput}), which is complementary to SAM 2 for practical deployment. 
    % Code and models will be released to benefit a wide range of efficient TAM applications. 
    
\end{itemize}

% \vspace{-2mm}
\section{Related Work}
% \vspace{-2mm}
\label{sec:formatting}
%We now review relevant works on segmentation, vision transformers, and efficient attention.

%-------------------------------------------------------------------------
% \subsection{Video Object Segmentation}
{\bf Video Object Segmentation (VOS)} is a fundamental task in computer vision, segments objects of interest from the background and tracks target objects in a video. 
%Many research works have been proposed in this community on video object segmentation. 
In the unsupervised setting~\citep{grundmann2010efficient,brox2010object,lee2011key,xu2012evaluation,fragkiadaki2012video,perazzi2012saliency,zhang2013video,li2013video,papazoglou2013fast,faktor2014video,wang2015saliency,taylor2015causal,perazzi2016benchmark}, VOS models segment salient objects without a reference mask. In the semi-supervised setting~\citep{pont20172017,xu2018youtube,oh2019video,bhat2020learning,robinson2020learning,li2022recurrent,yang2022decoupling,cheng2022xmem,zhang2023joint,wang2023look,wu2023scalable,cheng2024putting,yang2024scalable}, VOS requires tracking and segmenting objects based on a first-frame mask of target objects. For interactive video object segmentation (iVOS)~\citep{caelles20182018,heo2020interactive,cheng2021modular,homayounfar2021videoclick,yang2023track,cheng2023segment,rajivc2023segment,cheng2024putting,delatolas2024learning}, iVOS models perform object segmentation in videos (masklets) with user guidance, e.g., clicks, bounding boxes, scribbles. In SAM 2~\citep{ravi2024sam}. Semi-supervised VOS and iVOS have been extended to promptable visual segmentation (PVS), where the model can be interactively prompted with different types of inputs such as clicks, boxes, and masks on any frame in a video for segmenting and tracking a valid object.
%-------------------------------------------------------------------------

% \subsection{Vision Transformers}
\noindent {\bf Vision Transformers (ViTs)} have achieved huge success on various vision tasks including image classification~\citep{dosovitskiy2020image}, object detection~\citep{li2022exploring}, image segmentation~\cite{cheng2022masked,kirillov2023segment}, video classification~\citep{fan2021multiscale}, and video object segmentation~\citep{duke2021sstvos,yang2023track}. The original ViT family scales from the efficient ViT-Tiny up to ViT-Huge, with a plain, non-hierarchical architecture. There are also hierarchical vision transformers that combine transformers with hierarchical stage structure, such as Swin~\citep{liu2021swin}, MViT~\citep{fan2021multiscale,li2022mvitv2}, PViT~\citep{wang2021pyramid}, and Hiera~\citep{ryali2023hiera}. While being successful, hierarchical models are usually slower than their plain ViT counterparts for practical deployment~\citep{ryali2023hiera}. 
Combining ViT with convolutions~\citep{lecun1989backpropagation} has been explored for fast hybrid models such as MobileViT~\citep{mehta2021mobilevit}, LeViT~\citep{graham2021levit},  EfficientFormer\citep{li2022efficientformer}, Next-ViT\citep{li2022next}, Tiny-ViT\citep{wu2022tinyvit}, Castling-ViT\citep{you2023castling}, EfficientViT~\citep{liu2023efficientvit}, and MobileNetv4~\citep{qin2024mobilenetv4}. This line of progression towards building efficient ViTs is orthogonal to our
EfficientTAM work towards building efficient video object segmentation. Following SAM~\citep{kirillov2023segment} and EfficientSAMs~\citep{xiong2024efficientsam}, we are pursuing plain ViT backbones for efficient video object segmentation and track anything tasks.  
%The community has also shown increasing interest in efficient vision transformers; \citep{touvron2021training} presented smaller ViTs such as ViT-Small and ViT-Tiny for complementing ViT-Huge, ViT-Large, and ViT-Base in \citep{dosovitskiy2020image}. 

%-------------------------------------------------------------------------
% \subsection{Efficient Attention}
\noindent {\bf Efficient Attention.} The field has developed methods to reduce the quadratic cost of standard self-attention with respect to input sequence length~\cite{attention_is_all_you_need}. 
Local windowed attention has been applied in \cite{beltagy2020longformer,zaheer2020bigbird} for reducing the complexity of self-attention. In \cite{shen2018efficient,katharopoulos-et-al-2020}, a linear dot product approximation is proposed to linearize the softmax matrix in self-attention by heuristically separating keys and queries. In \cite{choromanski2020rethinking}, the Performer model uses random features to approximate self-attention, achieving linear time and memory cost. Nystr\"{o}mformer in \cite{xiong2021nystromformer} makes use of the Nystr\"{o}m method to approximate self-attention with a linear cost. Linformer \cite{wang2020linformer} shows that self-attention is low-rank, which can be approximated by learning linear projection matrices for the keys and values. The approach of~\citep{liu2023efficientvit,you2023castling} leverages the associative property of matrix multiplication for efficient attentions in vision transformers. This direction has shown success and has achieved decent performance on vision tasks. However, in preliminary experiments we found that these methods underperformed in a memory cross-attention module when adapted for efficiency improvement.

%-------------------------------------------------------------------------
% \subsection{Segment Anything Model}
\noindent {\bf Segment Anything Model.} SAM~\citep{kirillov2023segment} is a vision foundation model that can segment any object in an image using interactive prompts such as points and bounding boxes. SAM has demonstrated remarkable zero-shot transfer performance and high versatility for many vision tasks including a broad range of segmentation applications~\citep{chen2023semantic,cen2023sad,deng2023segment,chen2023sam}, in-painting~\citep{yu2023inpaint}, image restoration~\citep{jiang2023restore}, image editing~\citep{gao2023editanything}, image shadow removal~\citep{zhang2023deshadow}, medical image segmentation~\citep{ma2023segment}, camouflaged object detection~\citep{tang2023can}, transparent object detection~\citep{han2023segment}, concept-based explanation~\citep{sun2023explain}, semantic communication~\citep{tariq2023segment}, and object tracking~\citep{cheng2023segment,yang2023track}. The strong ability on image segmentation with flexible prompts motivates the extension of SAM for video object segmentation and track anything. Track Anything Model (TAM)~\citep{yang2023track} combines SAM and XMem~\cite{cheng2022xmem} for interactive video object tracking and segmentation with SAM for frame segmentation and XMem for tracking. SAM-Track~\citep{cheng2023segment} perform object tracking and segmentation in videos by combining SAM~\citep{kirillov2023segment}, DeAOT~\citep{yang2022decoupling}, and Grounding-Dino~\citep{liu2023grounding}. The latest SAM 2~\citep{ravi2024sam} extended SAM for video segmentation through a hierarchical image encoder for frame embeddings and a memory module that conditions current frame embeddings on past frames. Motivated by mobile app use-cases and computationally-constrained applications, recent works have reduced the computational cost of SAM, such as MobileSAM~\citep{zhang2023faster}, FastSAM~\citep{zhao2023fast}, and EfficientSAM~\citep{xiong2024efficientsam}.
The present paper focuses on improving the efficiency challenges of SAM 2 for practical deployment of video object segmentation and track anything.  
\section{Approach}
\subsection{Preliminaries}
\noindent \textbf{Segment Anything.} SAM~\citep{kirillov2023segment} contains a  ViT image encoder and a prompt-guided mask decoder. The encoder takes an image and outputs image embeddings. Then the decoder takes the image embeddings and a prompt, which allows cutting out any object from the background in an image. 
SAM is trained on an image dataset of over 1B masks.
% SAM is trained on SA-1B, an image dataset of over 1B masks from 11M images~\citep{kirillov2023segment}.

\noindent \textbf{Segment Anything 2.} The architecture of segment anything 2 (SAM 2)~\citep{ravi2024sam} largely follows SAM, which consists of a hierarchical image encoder, a prompt-guided lightweight mask decoder, and a new memory mechanism. SAM 2 uses a hierarchical image encoder, Hiera~\citep{ryali2023hiera}, to produce image embeddings for each frame. The stride 16 and 32 features from Stage 3 and 4 are used for the memory module. The stride 4 and 8 features from Stage 1 and Stage 2 are not used in the memory module but are fed to upsampling layers in the mask decoder for generating segmentation masks. For stable object tracking, SAM 2 employs a memory mechanism consisting of a lightweight memory encoder, a lightweight memory bank, and a memory attention module. It stores information from past frames and uses the memory attention module to perform cross-attention between the stored memory in the memory bank and current frame features, thereby understanding temporal dependencies in video.

The memory attention module consists of a stack of transformer blocks. Each block contains self-attention, cross-attention, and MLP. The first transformer block takes the image embedding from the current frame as an input. The core component of each transformer block, cross-attention, integrates the current frame embedding and the memory stored in memory bank to produce an embedding with temporal correspondence information. 
% Let us denote the memory bank as,
% \begin{equation*} 
%     \begin{split}
%     M_{b} = [m_{11}; \dots; m_{1N}; \dots, m_{T1}; \dots; m_{TN}] \in \R^{TN \times d_m},
%     \end{split}
% \end{equation*}
% where $T$ is the number of frames saved in the memory, $N$ is the number of memory tokens per frame, $d_m$ is the channel dimension. 
For memory tokens, it includes two parts, the spatial embedding tokens from memory encoder and the object-level pointer tokens from mask decoder. 
% Let us assume the resolution of spatial embedding tokens is $w \times h$, 
Let us assume the number of spatial tokens is $n$, the number of object-level pointer tokens is $P$, and $d_m$ is the channel dimension, memory tokens can be formulated as $
M_{b} = \begin{bmatrix} 
    M_s\in \R^{n\times d_m}\\
    M_p\in \R^{P\times d_m}
    \end{bmatrix}.
$
% \begin{equation} \label{equ:mbank}
%     \begin{split}
%     M_{b} = & [m_{11}; \dots; m_{1wh}; \dots, m_{1wh+P}; \\
%     & \vdots \dots \dots \vdots \dots \dots \vdots \dots \dots \vdots,\\
%     & m_{T1}; \dots; m_{Twh}; \dots, m_{T(wh+P)}]
%     \end{split}
% \end{equation}
% \[
% M_{b} = \begin{bmatrix} 
%     m_{11}; & \dots  & m_{1wh}; & \dots & m_{1wh};\\
%     \vdots & \dots & \vdots & \dots & \vdots\\
%     m_{T1}; & \dots  & m_{Twh}; &\dots & m_{Twh+P};
%     \end{bmatrix}
% \]

Let $L$ be the number of tokens and $d_q$ be the dimension of each token for input frame features after self-attention, $X \in \R^{L \times d_q}$. The input sequence $X \in \R^{L \times d_q}$ is linearly projected to input queries $Q \in \R^{L\times d}$, and the memory tokens, $M_{b} \in \R^{(n+P)\times d_m}$ are linearly projected to keys $K \in \R^{(n+P) \times d}$, and values $V \in \R^{(n+P) \times d}$ respectively, where $d$ is the embedding dimension of queries, keys, and values. The scaled dot-product cross attention mechanism applied on the queries $Q$, keys $K$, values $V$ can be formally written as, 
\begin{equation}\label{eq:crossattn}
     \textsf{C}(Q, K, V) =  \text{softmax}\Lleft\frac{QK^T}{\sqrt{d}}\Rright V,
\end{equation}
where the $\text{softmax}$ operation is applied row-wise. A single head cross attention is used in the memory module. In later discussion, we also consider keys and values as memory tokens for simplification. 
% The scaling $\sqrt{d}$ will be omitted in later discussion to simplify notation.

\noindent \textbf{Efficiency Bottleneck.} Despite the advantages of the hierarchical image encoder for multiscale frame feature extraction and cross-attention for integrating current frame features with stored memory, it poses the challenges for practical deployment of SAM 2. The inefficient SAM 2 (tiny) even shows comparable FPS to the base SAM 2, 47.2 FPS vs 43.8 FPS due to the hierarchical design of the image encoder and the use of hierarchical features, which also makes SAM 2 challenging to deploy on mobile devices. Moreover, the number of tokens in keys and values for performing cross-attention in the memory module are super long, e.g., $30K$. It leads to a large computation and memory cost when performing cross-attention, which becomes the efficiency bottleneck of the memory module for real-world deployment. 

\begin{figure*}[t]
    \centering
    \includegraphics[width=0.9\textwidth]{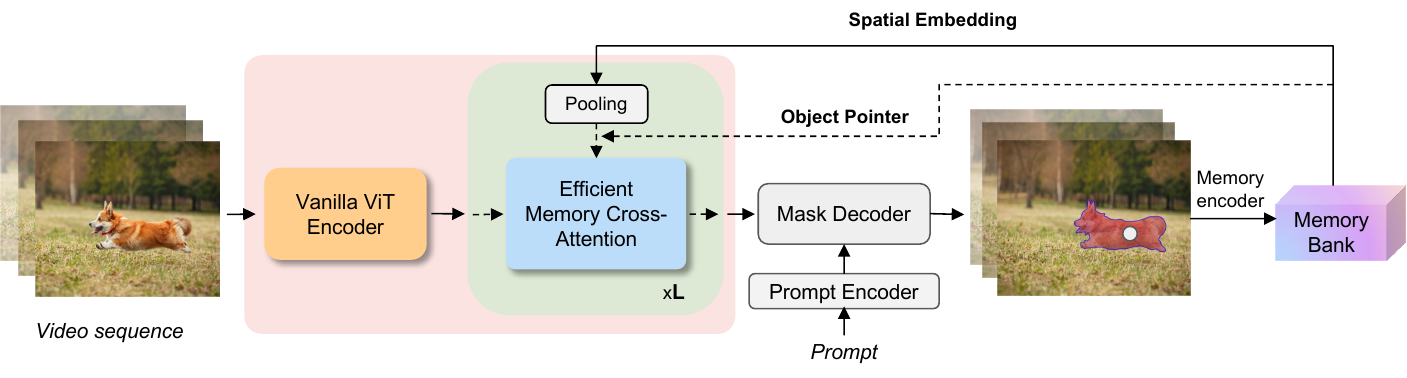}
    \caption{EfficientTAM architecture. Our proposed EfficientTAM takes a vanilla lightweight ViT image encoder for frame feature extraction. An efficient memory cross-attention is proposed to further improve the efficiency of EfficientTAM by leveraging the strong locality of memory spatial embeddings. EfficientTAM is fully trained on SA-1B (image) and SA-V (video) for unified image and video segmentation.}
    \label{fig:efficienttams}
\end{figure*}

\subsection{Efficient Video Object Segmentation and Track Anything} 
We now address the efficiency issue of SAM 2 for building efficient video object segmentation and track anything model, EfficientTAM. Motivated by the high quality segmentation performance of SAM and EfficientSAM, we revisit using plain, non-hierarchical lightweight image encoders such as ViT-Small/ViT-Tiny, for frame feature extraction. We found that the use of vanilla ViT for frame feature extraction makes EfficientTAM highly efficient and deployable on mobile devices. Further, we introduce an efficient memory module to reduce the computation and memory cost by proposing an efficient cross-attention operation. Based on these two designs, we build efficient video object segmentation and track anything model by largely following SAM2.  \cref{fig:efficienttams} illusrates an overview of our proposed EfficientTAM. 

\noindent \textbf{Efficient Image Encoder.} The image encoder's role is to produce feature embeddings for each high-resolution frame. We use a SAMI~\citep{xiong2024efficientsam} pretrained vanilla ViT image encoder~\citep{dosovitskiy2020image,touvron2021training} to extract frame features. Differing from the image encoder of SAM 2, our image encoder provides a single-scale feature map and no other features in the mask decoder are added to the upsampling layers during decoding for segmentation mask generation. We adopt the lightweight image encoders ViT-Small and ViT-Tiny with a $16\times 16$ patch size. Following \citep{li2022exploring}, we use $14\times 14$ non-overlapping windowed attention and 4 equally-spaced global attention blocks to efficiently extract features from high-resolution frames. Our image encoder outputs a single-scale feature embedding with a $16$x reduced resolution, which takes high-resolution (e.g., $1024\times 1024$) frames as input and transforms it into a dense embedding of downscaled size $64\times 64$. 

\begin{figure*}[t]
    \centering
    \begin{overpic}[width=1.0\linewidth]{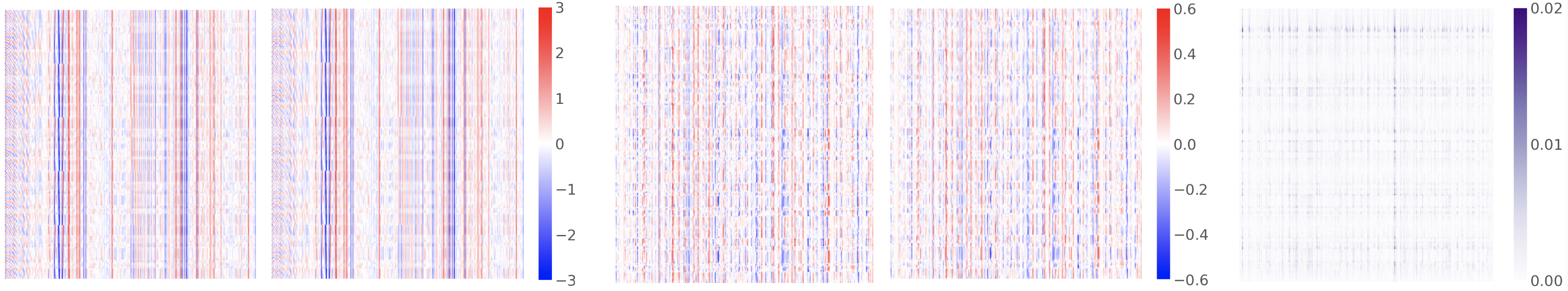}
\put (7,19) {\scriptsize{Keys}}
\put (22,19) {\scriptsize{Coarse Keys}}
\put (45,19) {\scriptsize{Values}}
\put (61,19) {\scriptsize{Coarse Values}}
\put (79,19) {\scriptsize{Absolute Cross-Attention Error }}
\end{overpic}
    \caption{An example to show strong locality of the Keys and Values in the cross-attention of the memory module. Keys and Values are a matrix of size $28700\times 256$. Cross-attention is a matrix of size $4096\times 256$. For simplicity of visualizing and comparison, we only draw the top matrix of size $320\times256$. We use a single averaged token to represent other tokens in the homogeneous window with a $2\times 2$ size, for Keys and Values to obtain coarse Keys and Values. At right, we visualize the difference between original cross-attention of \cref{eq:crossattn} and efficient cross-attention of \cref{eq:ecrossattn}; the relative error w.r.t original cross-attention is $0.03$ under Frobenius norm.}
    \label{fig:cross_attn}
% \vspace{-3mm}
\end{figure*}
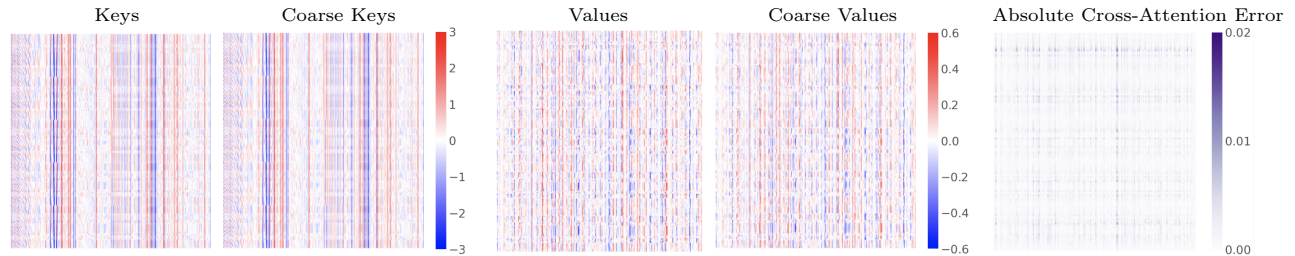

% The difference between original cross-attention of \cref{eq:crossattn} and efficient cross-attention of \cref{eq:ecrossattn} with coarse Keys and Values is marginal, visualized at the right. The relative error w.r.t. original cross-attention is $0.03$ under Frobenius norm. 

\noindent \textbf{Efficient Memory Module.} The memory module leverages information from previous frames to facilitate consistent object tracking. Cross-attention is a major efficiency bottleneck of the memory module in SAM 2~\citep{ravi2024sam} due to its long memory token sequence. We now discuss how exploiting the underlying structure of memory tokens --- local smoothness (strong locality) within spatial memory tokens --- can yield a more efficient cross-attention. 

Consider two consecutive memory spatial tokens, $k_i$ and $k_{i+1}$, local smoothness  implies that $||k_i - k_{i+1}||^2_2 \leq \frac{c_K}{n^2}$, for $i = 1, \dots, n - 1$, where $c_K$ is a positive constant. 
This suggests that given a sufficient small local window, $l_w \times l_h$, using a single token to represent other tokens in the homogeneous window may provide a coarser representation of the full set of memory spatial tokens $K_s$ as $\Tilde{K}_s$. We can construct a good surrogate of $K_s$ with the same size, $\Bar{K}_s$, from $\Tilde{K}_s$ by repeating the single token in each window $l_w\times l_h$ times. Under the smoothness assumption, $\Bar{K}_s$ will not be far from $K_s$. Empirically, we observed that a coarser representation of spatial memory tokens is good surrogate of the full spatial memory tokens. \cref{fig:cross_attn} confirms the coarser representation of input keys and values are close to the original keys and values of cross-attention in the memory module. 

Utilizing highly correlated neighboring tokens in cross-attention, we perform average pooling to efficiently compute a coarser representation for keys $K$ and values $V$ in our model. For input spatial tokens $K_s = [k_{11}, \dots, k_{1h}; \dots; k_{w1}, \dots, k_{wh}]$ where $w \times h$ is the resolution size, we divide the $n = w\times h$ tokens into $k = \Tilde{w}\times \Tilde{h}$ rectangular pooling regions and compute the average token of each region. For simplicity, we assume $w$ is divisible by $\Tilde{w}$ and $h$ is divisible by $\Tilde{h}$. Denote $l_w = \frac{w}{\Tilde{w}}, l_h = \frac{h}{\Tilde{h}}$. $\Tilde{K}_s$ and $\Tilde{V}_s$ can be computed by averaging each region as,   
\begin{align}\label{eq:keys-values}
\small
\Tilde{k}_{ij} & = \sum_{p = i \times l_w + 1}^{(i + 1) \times l_w}\sum_{q = j \times l_h + 1}^{(j + 1) \times l_h} \frac{k_{pq}}{l_w \times l_h}, \nonumber \\
\Tilde{v}_{ij} & = \sum_{p = i \times l_w + 1}^{(i + 1) \times l_w}\sum_{q = j \times l_h + 1}^{(j + 1) \times l_h} \frac{v_{pq}}{l_w \times l_h}, 
\end{align}
where $i = 1, \cdots, \Tilde{w}, j=1, \cdots, \Tilde{h}$. This token-pooling scheme requires a single scan
of the tokens leading to an efficient coarse token generation. We find that using averaging pooling with window size, $2 \times 2$, is sufficient to ensure a good approximation for spatial memory tokens. 

Assume $\Tilde{K}_s$ is a coarser representation of memory spatial keys, $K_s$, we can construct a good surrogate of $K_s \in \R^{n \times d}$ with the same size, $\Bar{K}_s \in \R^{n \times d}$ from $\Tilde{K}_s \in \R^{\Tilde{w}\Tilde{h} \times d}$ by stacking each $\Tilde{k}_i, i = 1, \dots, \Tilde{w}\Tilde{h}$, $l_w\times l_h$ times, which can be written as,
\begin{align*}
    \Bar{K}_s = [\underbrace{\Tilde{k}_1; \dots; \Tilde{k}_1}_{l_w\times l_h}; \underbrace{\Tilde{k}_2; \dots; \Tilde{k}_2}_{l_w\times l_h}; \dots; \underbrace{\Tilde{k}_{\Tilde{w}\Tilde{h}}; \dots; \Tilde{k}_{\Tilde{w}\Tilde{h}}}_{l_w\times l_h}]
\end{align*}
Similarly, we stack each $\Tilde{v}_i, i =1, \dots, \Tilde{w}\Tilde{h}$, $l_w\times l_h$ times to construct $\Bar{V}_s  \in \R^{n \times d}$ as a good surrogate of values, $V_s \in \R^{n\times d}$, which can be written as,
\begin{align*}
    \small 
    \Bar{V}_s= [\underbrace{\Tilde{v}_1; \dots; \Tilde{v}_1}_{l_w\times l_h}; \underbrace{\Tilde{v}_2; \dots; \Tilde{v}_2}_{l_w \times l_h}; \dots; \underbrace{\Tilde{v}_{\Tilde{w}\Tilde{h}}; \dots; \Tilde{v}_{\Tilde{w}\Tilde{h}}}_{l_w \times l_h}]
\end{align*}
% With the stacked $\Bar{K}_s \in \R^{n \times d}$, $\Bar{V}_s \in \R^{n \times d}$,
Then we concatenate this coarse spatial tokens with object pointer tokens, $\Bar{K} = [\Bar{K}_s; K_p]\in \R^{(n+P)\times d}$ and $\Bar{V} = [\Bar{V}_s; K_p]\in \R^{(n+P)\times d}$, for a good surrogate of original memory tokens, $K$ and $V$. For the coarse memory tokens, $\bar{K}$ and $\bar{V}$, we have,
% \begin{align}\label{eq:replace}
%     \text{softmax}\Lleft\frac{Q\Bar{K}^{T}}{\sqrt{d}}\Rright\Bar{V} = \text{softmax}\Lleft\frac{Q\Tilde{K}^{T}}{\sqrt{d}} + \ln{(l_w\times l_h)}\Rright\Tilde{V},
% \end{align}
\begin{align}\label{eq:replace}
    \text{softmax}\Lleft\frac{Q\Bar{K}^{T}}{\sqrt{d}}\Rright\Bar{V} = \text{softmax}\Lleft A \Rright\Tilde{V},
\end{align}
where $A = [\frac{Q\Tilde{K}_s^{T}}{\sqrt{d}} + \ln{(l_w\times l_h)}, \frac{QK_p^{T}}{\sqrt{d}}] \in \R^{L\times (\Tilde{w}\Tilde{h} + P)}$, $\Tilde{V} = [\Tilde{V}_s; V_p] \in \R^{(\Tilde{w}\Tilde{h}+P)\times d}$. We provide a proof of \cref{eq:replace} in the appendix. 
Since $\Bar{K}$ and $\Bar{V}$ are good surrogate of $K$ and $V$ respectively, we obtain a good surrogate of the original cross-attention, $\text{softmax}\Lleft\frac{QK^T}{\sqrt{d}}\Rright V$ in \cref{eq:crossattn},
\begin{equation}\label{eq:scrossattn}
     \Bar{\textsf{C}}(Q, K, V) =  \text{softmax}\Lleft\frac{Q\bar{K}^T}{\sqrt{d}}\Rright \bar{V}.
\end{equation}
With \cref{eq:replace}, we have an efficient version of cross-attention, 
\begin{equation}\label{eq:ecrossattn}
     \Bar{\textsf{C}}(Q, K, V) = \text{softmax}(A)\Tilde{V}.
\end{equation}
% \begin{equation}\label{eq:ecrossattn}
%      \Bar{\textsf{C}}(Q, K, V) = \text{softmax}\Lleft\frac{Q\Tilde{K}^{T}}{\sqrt{d}} + \ln{(l_w\times l_h)}\Rright\Tilde{V},
% \end{equation}
\noindent\textbf{Link to efficient cross-attention variants.} Interestingly, we can find some cross-attention variants based on our proposed efficient cross-attention in  \cref{eq:ecrossattn}. We notice there is a constant for balancing the attention score between coarse spatial tokens and object pointer tokens, avoiding reducing the attention to spatial tokens after pooling. If we remove this constant, it can lead to a linformer variant using averaging pooling to replace the learnable projection. Instead of removing the constant, we add it to keys for regularizing the attention between coarse spatial tokens and object pointer tokens in \cref{eq:acrossattn}, for obtaining another variant. 

\begin{equation}\label{eq:acrossattn}
     \Tilde{\textsf{C}}(Q, K, V) = \text{softmax}\Lleft\frac{Q\Tilde{K}^{T}}{\sqrt{d}} \Rright\Tilde{V},
\end{equation}
where $\Tilde{K} = [\Tilde{K}_s + \ln{(l_w\times l_h)}, K_p]\in\R^{(\Tilde{w}\Tilde{h}+P)\times d}$.

It is feasible to achieve a good surrogate of the original cross-attention because spatial memory embeddings have strong locality. 
% since the neighboring tokens are similar. 
Our efficient cross-attention is close to the original cross-attention, visualized in \cref{fig:cross_attn}. 

\section{Experiments}
\subsection{Experimental Setting}
\noindent \textbf{Pretraining.} The SA-1B dataset  consists of 11M diverse, high resolution images with 1.1B high-quality segmentation masks. Similar to~\citep{ravi2024sam}, we pretrain our EfficientTAM without memory components on SA-1B dataset~\citep{kirillov2023segment} for 90k steps. Our ViT image encoder is initialized from pre-trained ViTs~\citep{xiong2024efficientsam}
. We use the AdamW optimizer ~\citep{loshchilov2017decoupled} with a momentum, ($\beta_1 = 0.9$, $\beta_2 = 0.999$), a global batch size of 256, and a initial learning rate of $4e-4$. The learning rate is decayed by a reciprocal square root learning rate schedule~\citep{zhai2022scaling} with 1k iterations linear warmup and 5k iterations linear cooldown. We set weight decay to 0.1. We do not apply drop path for our image encoder. Layer-wise decay~\citep{clark2020electra} is set to 0.8. We apply horizontal flip augmentation and resize the input image resolution to $1024\times 1024$. We restrict our training to $64$ masks per image. Our models are pre-trained on 256 A100 GPUs with 80GB GPU memory with a linear combination of focal and dice loss for mask prediction (e.g., a ratio of 20:1). Bfloat16 is used during the training.

\noindent \textbf{Full Training Datasets.} Following~\citep{ravi2024sam}, we train our EfficientTAM including memory components on SA-V dataset~\citep{ravi2024sam} and a 10\% subset of SA-1B~\citep{kirillov2023segment}. SA-V is a large-scale and diverse video segmentation dataset, including 51K videos captured across 47 countries and 600K mask annotations covering whole objects and parts. SA-V video resolution ranges from 240p to 4K and duration ranges from 4 seconds to 138 seconds. Unlike SAM 2, we do not use other open-source datasets or internal datasets during our training for a fair comparison with baselines. 

\noindent \textbf{Full Training Implementation Details.} Similar to  ~\citep{ravi2024sam}, we train our EfficientTAM for 300k steps after pretraining. We use the AdamW optimizer ~\citep{loshchilov2017decoupled} with a momentum, ($\beta_1 = 0.9$, $\beta_2 = 0.999$), a batch size of 256, and a initial learning rate of $6e-5$ for image encoder and $3e-4$ for other components of the model. The learning rate is decayed by a cosine schedule with 15k iterations linear warmup. We set weight decay to 0.1. We do not apply drop path for our image encoder. Layer-wise decay~\citep{clark2020electra} is set to 0.8. We apply horizontal flip image augmentation and resize the input image resolution to $1024\times 1024$. For video, we apply horizontal flip augmentation, affine transformation with degree $25$ and shear $20$, color jittering with brightness $0.1$, contrast $0.03$, saturation $0.03$, gray scale augmentation with a probability of $0.05$, We restrict our training to $64$ masks per image and $3$ masks per frame for video. Our models are trained on 256 A100-80G GPUs with a linear combination of focal and dice losses for mask prediction, mean-absolution-error loss for IoU prediction, and cross-entropy loss for object prediction. The ratio for the linear combination loss is 20:1:1:1. Bfloat16 is used for training.

\noindent \textbf{Downstream Tasks/Datasets/Models.} \underline{\textit{Tasks and Datasets.}} We consider zero-shot video tasks including promptable video segmentation and semi-supervised video object segmentation, and zero-shot image tasks to demonstrate the competing capabilities of EfficientTAM on image and video segmentation.
For zero-shot image tasks, we evaluate EfficientTAM on 37 datasets including 23 datasets of SA-23~\citep{kirillov2023segment} and 14 video datasets introduced in~\citep{ravi2024sam}. For zero-shot video tasks, we evaluate our EfficientTAM on 9 densely annotated datasets for promptable video segmentation. We use 17 video datasets to evaluate zero-shot accuracy under interactive semi-supervised VOS setting using different prompts. For the standard semi-supervised VOS setting where a ground-truth mask on the first frame is provided, MOSE~\citep{ding2023mose}, DAVIS2017~\citep{pont20172017}, LVOS~\citep{hong2024lvos}, SA-V~\citep{ravi2024sam}, and YTVOS~\citep{xu2018youtube} are used to measure the VOS accuracy. We refer readers to~\citep{kirillov2023segment,ravi2024sam} for the details of these datasets.
\underline{\textit{Models.}} We use our EfficientTAM for zero-shot image and video tasks.

\noindent\textbf{Baselines and Evaluation Metrics.}
\underline{\textit{Baselines.}} For the standard semi-supervised VOS task, where the first-frame mask is provided, we compare the performance of our EfficientTAM with SAM 2\citep{ravi2024sam}, Cutie-base\citep{cheng2024putting}, DEVA~\citep{cheng2023tracking}, XMem~\citep{cheng2022xmem}, etc.
For the zero-shot promptable video segmentation task and the interactive semi-supervised video object segmentation task using different prompts, we compare our method with SAM2~\citep{ravi2024sam}, SAM+XMem++~\citep{ravi2024sam}, and SAM+Cutie~\citep{ravi2024sam}. For zero-shot image segmentation task, we compare with SAM~\citep{kirillov2023segment} and SAM2~\citep{ravi2024sam}. Note that we use the opensource version of SAM 2 (without training on MOSE/LVOS/YTVOS) for comparison. We also acknowledge the very recent release of SAM 2.1 trained with long memory contexts. 
\underline{\textit{Evaluation Metrics.}} We evaluate our method and all baselines using the accuracy metrics of the combined $\mathcal{J}$(region similarity)\&$\mathcal{F}$(contour accuracy),  for zero-shot video segmentation tasks; mIoU (mean intersection over union) for zero-shot image segmentation tasks. For efficiency metrics, we compare the number of model parameters or inference throughput on GPU (e.g, A100) and latency on mobile devices (e.g., iPhone 15 Pro Max). We follow SAM 2~\citep{ravi2024sam} to report metrics. When providing main results on MOSE, LVOS and YTVOS, we submit to their benchmarking servers to evaluate on, \textit{MOSE val}, \textit{LVOS val}, and \textit{YTVOS2019 val}, for final performance. For ablation studies, we evaluate on a MOSE development set, \textit{MOSE dev} with 200 randomly-sampled videos from the MOSE training split~\citep{ravi2024sam}.

\subsection{Main Results}
% Please add the following required packages to your document preamble:
% \usepackage{multirow}
% Please add the following required packages to your document preamble:
% \usepackage{multirow}
\begin{table*}[t]
% \vspace{-3mm}
\centering
\resizebox{0.9\linewidth}{!}{
\begin{tabular}{c|cccc|c|c|c|c}
\hline
\multirow{2}{*}{Method}                                   & \multicolumn{4}{c|}{$\mathcal{J}$\&$\mathcal{F}$}                                                                                                                                                         & $\mathcal{G}$                                            & \multirow{2}{*}{\begin{tabular}[c]{@{}c@{}}\\Parameters\\ (M)\end{tabular}} & FPS   & Latency (ms) \\ \cline{2-6} \cline{8-9} 
                                                          & \begin{tabular}[c]{@{}c@{}}MOSE \\ val\end{tabular} & \begin{tabular}[c]{@{}c@{}}DAVIS\\ 2017 val\end{tabular} & \begin{tabular}[c]{@{}c@{}}LVOS\\ val\end{tabular} & \begin{tabular}[c]{@{}c@{}}SA-V\\ test\end{tabular} & \begin{tabular}[c]{@{}c@{}}YTVOS\\ 2019 val\end{tabular} &                                                                           & A100  & iPhone15     \\ \hline
STCN~\citep{cheng2021rethinking}                                                      & 52.5                                                & 85.4                                                     & -                                                  & 57.3                                                & 82.7                                                     &   54                                                                        & 62.8  & -            \\
% R50-AOT-L~\citep{yang2021associating}                                                 & 58.4                                                & 84.9                                                     & -                                                  & 56.7                                                & 85.3                                                     & 15                                                                          & 6.4   & -            \\
% DeAOT-R50~\citep{yang2022decoupling}                                                 & 64.1                                                & 86.0                                                     & -                                                  & 59.9                                                & 85.3                                                     &                                                              20             & 11.7  & -            \\
RDE~\citep{li2022recurrent}                                                       & 46.8                                                & 84.2                                                     & -                                                  & 48.4                                                & 81.9                                                     &      64                                                                     & 88.8  & -            \\
XMem~\citep{cheng2022xmem}                                                      & 59.6                                                & 86.0                                                     & -                                                  & 60.1                                                & 85.6                                                     &     62                                                                      & 61.2  & -            \\
% SimVOS-B~\citep{wu2023scalable}                                                 & -                                                   & 88.0                                                     & -                                                  & 41.2                                                & 84.2                                                     &                                                                           & 3.3   & -            \\
% JointFormer~\citep{zhang2023joint}                                               & -                                                   & 90.1                                                     & -                                                  & -                                                   & 87.4                                                     &                                                                           & 3.0   & -            \\
% ISVOS~\citep{wang2023look}                                                     & -                                                   & 88.2                                                     & -                                                  & -                                                   & 86.3                                                     &                                                                           & 5.8   & -            \\
DEVA~\citep{cheng2023tracking}                                                      & 66.0                                                & 87.0                                                     & 55.9                                               & 53.8                                                & 85.4                                                     &      69                                                                     & 65.2  & -            \\
Cutie-base~\citep{cheng2024putting}                                                & 69.9                                                & 87.9                                                     & 66.0                                               & 61.6                                                & 87.0                                                     & 35                                                                        & 65    & -            \\
Cutie-base+~\citep{cheng2024putting}                                               & 71.7                                                & 88.1                                                     & -                                                  & 62.3                                                & 87.5                                                     & 35                                                                        & 57.2  & -            \\
SAM 2~\citep{ravi2024sam}                                                     & 72.8                                                & 88.9                                                     &             76.2                                       &         74.7                                        & 87.9                                                     & 81                                                                        & 43.8  & -            \\ \hline
\rowcolor{gray} EfficientTAM-Ti/2 (ours) & 68.4                                                & 88.4                                                     & 66.1                                               & 70.8                                                & 87.1                                                     & 18                                                                        & 109.4 & 261.4        \\
\rowcolor{gray} EfficientTAM-Ti (ours)   & 69.3                                                & 89.1                                                     & 69.6                                               & 70.7                                                & 86.7                                                     & 18                                                                        & 96.2  & 840.5        \\
\rowcolor{gray} EfficientTAM-S/2 (ours)  & 70.8                                                & 88.6                                                     & 72.1                                               & 74.0                                                & 87.2                                                     & 34                                                                        & 109.4 & 450          \\
\rowcolor{gray} EfficientTAM-S (ours)    & 71.4                                                & 89.2                                                     & 73.4                                               & 74.5                                                & 87.2                                                     & 34                                                                        & 85.0  & 1010.8       \\ \hline
\end{tabular}}
\caption{Standard semi-supervised video object segmentation results across video object segmentation benchmarks.}
\label{tab:vos}
\end{table*}
\textbf{Standard Semi-Supervised Video Object Segmentation.} Semi-supervised video object segmentation is the process of object segmentation and tracking in a video based on a ground-truth mask on the first frame. We follow SAM 2~\citep{ravi2024sam} and report accuracy of our methods on this standard semi-supervised video object segmentation task. We also report latency on a single A100 GPU with a batch size of 1. We evaluate EfficientTAMs with different image encoders, ViT-Tiny and ViT-Small, and memory modules, original memory block and efficient memory block with a $2\times2$ window pooling for a trade-off between efficiency and accuracy. EfficientTAM-S denotes EfficientTAM using a ViT-Small image encoder and the original memory block, and EfficientTAM-S/2 denotes EfficientTAM with a ViT-Small image encoder and efficiency memory block with a $2\times 2$ window pooling. \cref{tab:vos} compares our EfficientTAM with VOS baselines including SAM 2~\citep{ravi2024sam}, Cutie-base~\citep{cheng2024putting}, and XMem~\citep{cheng2022xmem}. On SA-V test, our EfficientTAM-S achieves 74.5 $\mathcal{J}$\&$\mathcal{F}$, outperforming Cutie-base, Cutie-base+, and XMem by 12.2, 12.9, and 14.4, respectively. On long-term video object segmentation benchmark, LVOS, we can also see that Our EfficientTAM-S outperform Cutie-base and XMem by a large margin. Notice that our EfficientTAM-S only underperforms SAM 2 by $<2$ $\mathcal{J}$\&$\mathcal{F}$ or $\mathcal{G}$ across 5 video benchmarks with $\sim$2x speedup and $\sim$2.4x fewer parameters. Further, EfficientTAM with efficient memory attention performs slightly worse than the one with original memory attention, but with much speedup, especially on mobile devices, $>$2x reduced latency on iPhone 15. For example, EfficientSAM-S achieves 74.5 $\mathcal{J}$\&$\mathcal{F}$ on SA-V test with 1010.8 ms running time per frame on iPhone 15. EfficientSAM-S/2 with efficient cross-memory attention obtain 74.0 $\mathcal{J}$\&$\mathcal{F}$ with only 450 ms. These results show the extraordinary benefits of EfficientTAMs for semi-supervised video object segmentation and validate the advantages of our methods for practical deployment.

\begin{figure*}[t]
    \centering
    \begin{overpic}[width=0.4\linewidth]{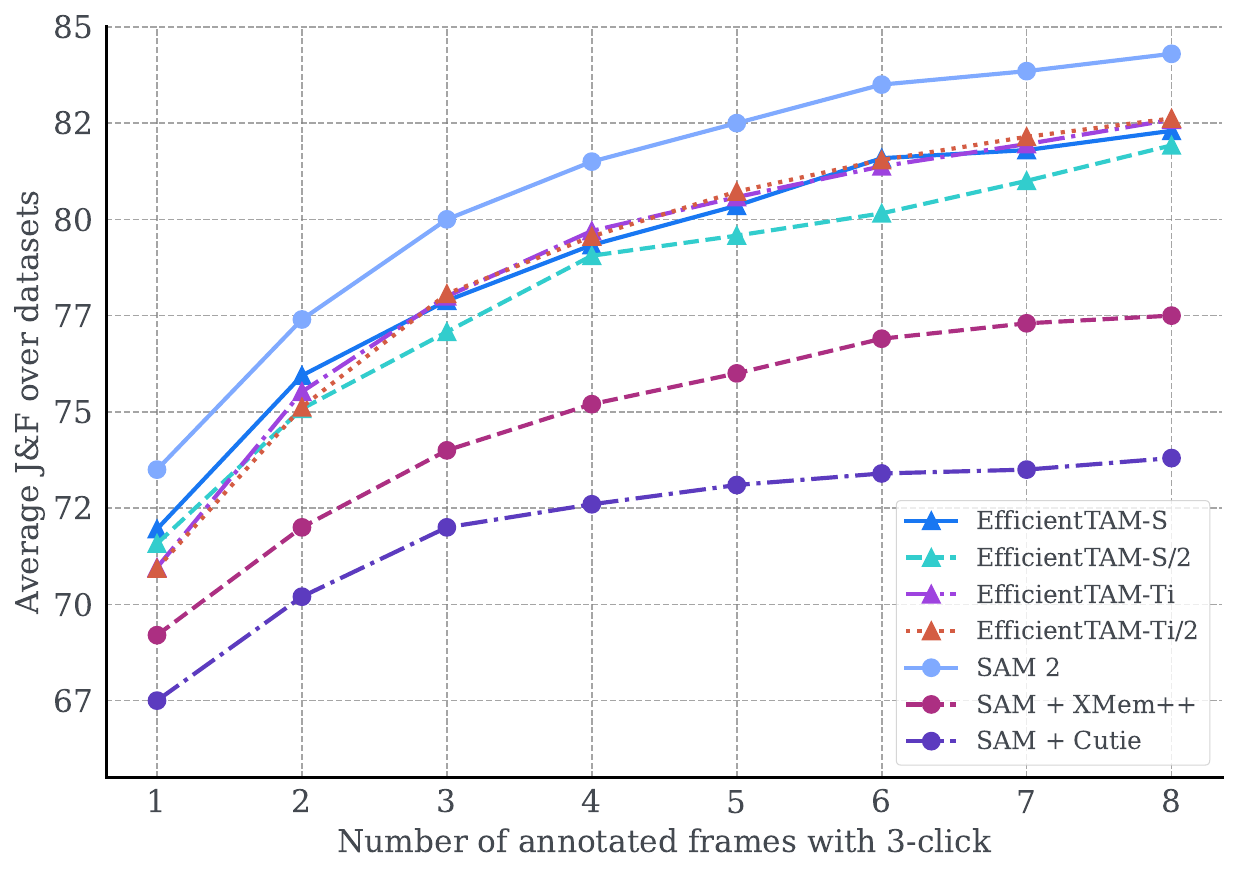}
    \end{overpic}
    \begin{overpic}[width=0.4\linewidth]{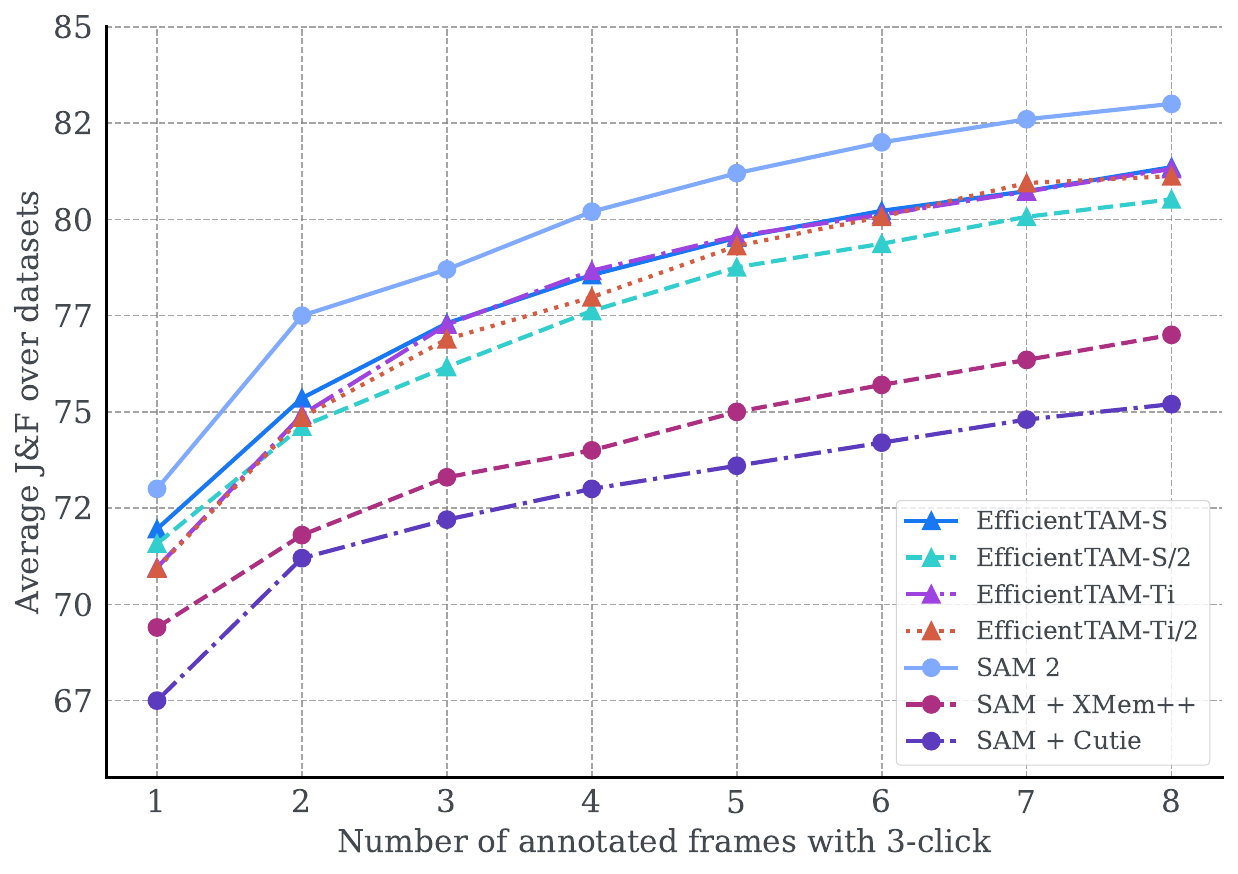}
    \end{overpic}
    \caption{Promptable video segmentation results across 9 video segmentation datasets under interactive offline (left) and online (right) evaluation settings. The average $\mathcal{J}$\&$\mathcal{F}$ over $1, \dots, 8$ interacted frames is reported.}
    \label{fig:pvs}
\end{figure*}

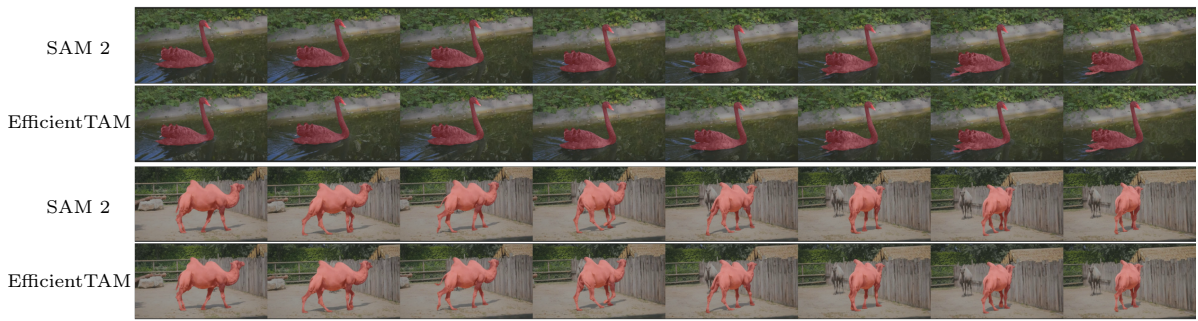
\begin{figure*}[h]
\centering
% \vspace{5pt}
\begin{overpic}[width=0.85\linewidth]{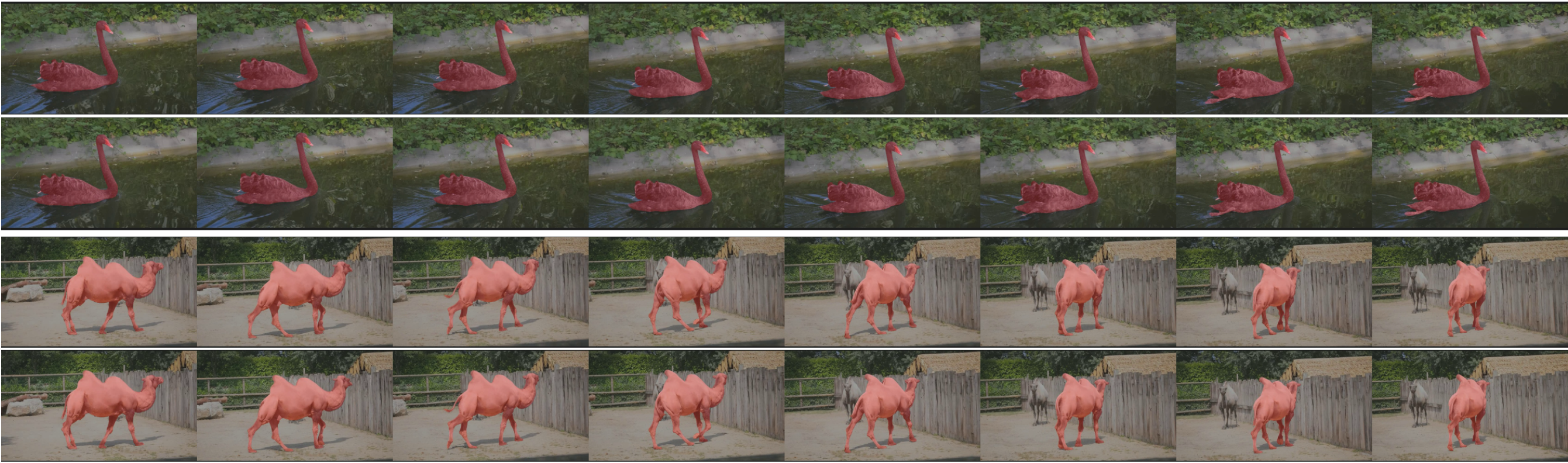}
\put (-8.3,25) {\scriptsize{SAM 2}}
\put (-12.0,18) {\scriptsize{EfficientTAM}}
\put (-8.3,10) {\scriptsize{SAM 2}}
\put (-12.0,3) {\scriptsize{EfficientTAM}}
\end{overpic}
\caption{Visualization results on video segmentation and tracking with SAM 2, and our EfficientTAM model. We sampled a subset of frames for visualization. The segmented objects, e.g., the goose and the camel, are colored in red. }
\label{fig:visual_vost}
\end{figure*}

\noindent \textbf{Promptable Video Segmentation.} Similar to SAM 2~\citep{ravi2024sam}, we evaluate promptable video segmentation using two settings, offline evaluation and online evaluation. For offline evaluation, we make multiple passes through a video to annotate frames  w.r.t. the largest model error. For online evaluation, we make a single pass through the video to annotate frames. 3 clicks per frame are used for the evaluations on 9 densely annotated video datasets including 
EndoVis, ESD, LVOSv2, LV-VIS, UVO, VOST, PUMaVOS, Virtual KITTI 2, and VIPSeg. Average $\mathcal{J}$\&$\mathcal{F}$ accuracy over $1, \dots, 8$ interacted frames is reported. \cref{fig:pvs} shows the comparison between our method and strong baselines including SAM 2, SAM + XMem++, and SAM + Cutie. EfficientTAM outperforms SAM + XMem++ and SAM + Cutie for both evaluation settings. EfficientTAM also reduces the gap between SAM 2 for offline and online settings. Specifically, with 8 annotated frames with 3-click, EfficientTAM-S and EfficientTAM-S/2 achieve $\sim$ 82 $\mathcal{J}$\&$\mathcal{F}$ in average for offline evaluation setting and $\sim$ 81 $\mathcal{J}$\&$\mathcal{F}$ in average for online evaluation, outperforming SAM + XMem++, and SAM + Cutie by $>$3 $\mathcal{J}$\&$\mathcal{F}$ and reducing the gap of SAM 2. This set of experiments further validate the effectiveness of our EfficientTAM on promptable video segmentation. 

\begin{table}[t]
    \centering
    \resizebox{0.8\linewidth}{!}{
    \begin{tabular}{c|ccccc}
    \hline
    Method           & 1-click & 3-click & 5-click & bounding box & ground-truth mask \\ \hline
    SAM+XMem++       & 56.9    & 68.4    & 70.6    & 67.6         & 72.7              \\
    SAM+Cutie        & 56.7    & 70.1    & 72.2    & 69.4         & 74.1              \\
    SAM 2            & 64.3    & 73.2    & 75.4    & 72.9         & 77.6              \\ \hline
    \rowcolor{gray} EfficientTAM-S/2 & 60.5    & 72.8    & 75.4    & 71.2         & 76.8              \\
    \rowcolor{gray} EfficientTAM-S   & 63      & 74.1    & 75.7    & 73.2         & 77.8              \\ \hline
    \end{tabular}}
    \caption{Interactive semi-supervised video object segmentation results with different prompts. We report averaged $\mathcal{J}$\&$\mathcal{F}$ zero-shot accuracy across 17 video datasets for each type of prompt.}
    \label{tab:interactive}
\end{table}
\noindent \textbf{Interactive Semi-Supervised Video Object Segmentation.} We also evaluate our method on the interactive semi-supervised video object segmentation task with click, box, or mask prompts provided only on the first frame by following SAM 2. In \cref{tab:interactive}, we report the average $\mathcal{J}$\&$\mathcal{F}$ accuracy over 17 video datasets for each type of prompt. We observe that EfficientTAM outperforms SAM + XMem++, and SAM + Cutie with different input prompts. We also notice the reduced gap between EfficientTAM and SAM 2. With 1 click, our EfficientTAM-S obtain 63 $\mathcal{J}$\&$\mathcal{F}$ accuracy, with a 6 $\mathcal{J}$\&$\mathcal{F}$ gain over SAM + XMem++ and SAM + Cutie and a slight loss, 1.3 $\mathcal{J}$\&$\mathcal{F}$ comparing to SAM 2. In summary, EfficientTAM performs favorably on the interactive semi-supervised VOS task using different prompts. 

\begin{table}[t]
\centering
\resizebox{0.75\linewidth}{!}{
\begin{tabular}{c|cccc}
\hline
Model             & SA-23 All   & SA-23 Image & SA-23 Video & 14 new Video \\ \hline
SAM (ViT-B)       & 55.9 (80.9) & 57.4 (81.3) & 54.0 (80.4) & 54.5 (82.6)  \\
SAM (ViT-H)       & 58.1 (81.3) & 60.8 (82.1) & 54.5 (80.3) & 59.1 (83.4)  \\
HQ-SAM (ViT-B)    & 53.9 (72.1) & 56.3 (73.9) & 50.7 (69.9) & 54.5 (75.0)  \\
HQ-SAM (ViT-H)    & 59.1 (79.8) & 61.8 (80.5) & 55.7 (78.9) & 58.9 (81.6)  \\
SAM 2              & 61.9 (83.6) & 63.2 (83.8) & 60.3 (83.3) & 69.9 (85.9)  \\ \hline
\rowcolor{gray} EfficientTAM-Ti/2 (ours) & 58.6 (82.5) & 59.6 (82.8) & 57.4 (82.1) & 63.4 (84.9)  \\
\rowcolor{gray} EfficientTAM-Ti (ours)   & 58.2 (82.6) & 59.5 (82.9) & 56.5 (82.1) & 62.7 (85.0)  \\
\rowcolor{gray} EfficientTAM-S/2 (ours)  & 60.5 (82.9) & 61.6 (83.2) & 59.1 (82.4) & 67.8 (85.4)  \\
\rowcolor{gray} EfficientTAM-S (ours)   & 60.7 (83.0) & 61.7 (83.3) & 59.5 (82.6) & 67.7 (85.4)  \\ \hline
\end{tabular}}
\caption{Segment anything results on SA-23 benchmark~\citep{kirillov2023segment} and 14 new video benchmark~\citep{ravi2024sam}. The average 1-click (5-click) mIoU is reported.}
\label{tab:sa23}
\end{table}
\noindent \textbf{Segment Anything on Images.} We now evaluate our model for the segment anything task on images. In Table \cref{tab:sa23}, we report 1-click and 5-click mIoU accuracy on both SA-23 benchmark, plus the new benchmark introduced in SAM 2~\citep{ravi2024sam} with 14 video datasets from video domain. We compare our EfficientTAMs with SAM (ViT-H) and HQ-SAM (ViT-H). Our EfficientTAM-S obtains a 2.6 mIoU improvement over SAM (ViT-H) and 1.6 mIoU improvement over HQ-SAM (ViT-H) on 1-click accuracy. For 5-click, we observe consistent improvement over SAM (ViT-H) and HQ-SAM (ViT-H). We also notice a significant improvement on the video benchmarks of SA-23 and the one with 14 new videos. This indicates our EfficientTAMs are strong for both image and video segmentation.

\noindent \textbf{Qualitative Evaluation.} 
\cref{fig:visual_vost} shows two video examples. We compare EfficientTAM and SAM 2 with a mask in the first frame prompted. We find that our EfficientTAM can generate high-quality masklet for the target object as SAM 2. More video examples are in the appendix. These results suggest that our EfficientTAMs have similar abilities to SAM 2, while EfficientTAM is more efficient. 

\vspace{-1mm}
\subsection{Ablation Studies}
\textbf{Impact of the object pointer tokens.} We study the effect of the object pointer tokens when performing cross-attention in the memory module. We ablate the cross-attention with or without the object pointer tokens. We find that object pointers significantly improve the performance on SA-V test dataset, 74.5 vs 72.1 $\mathcal{J}$\&$\mathcal{F}$, consistent with SAM 2~\citep{ravi2024sam}. This demonstrates that object pointer tokens need to be cross-attended with spatial tokens from the memory bank.

\noindent \textbf{Structure of memory tokens.} We ablate the impact of memory tokens for efficient cross-attention in the memory module. In our efficient cross-attention, we leverage the locality of memory spatial tokens for a coarser representation, and we concatenate the coarser embedding with object pointer tokens. We observe that naively pooling the entire memory tokens instead of only the spatial tokens yields a large performance drop, 2.3 $\mathcal{J}$\&$\mathcal{F}$ on SA-V test.

\noindent \textbf{Impact of window size.} We perform an averaging pooling for a good surrogate in \cref{eq:ecrossattn}. We experiment with window sizes $2\times 2$ and $4 \times 4$. We find increasing the window from $2\times 2$ to $4\times 4$ for efficient cross-attention will lead to $\sim$ 1 $\mathcal{J}$\&$\mathcal{F}$ accuracy drop with marginal speed improvement. Therefore, we use window size $2\times 2$ to achieve a trade-off between accuracy and efficiency.

\noindent \textbf{Linear cross-attention.} We explore adapting one representative efficient attention method such as linear attention~\citep{choromanski2020rethinking,cai2023efficientvit,you2023castling} by leveraging the associative property of matrix multiplication. We find that linear attention using associative property of matrix multiplication leads to significant performance drop, $>10$ $\mathcal{J}$\&$\mathcal{F}$ accuracy on SA-V test, comparing to our proposed efficient cross-attention. Therefore, leveraging the underlying token structure for efficient cross-attention is more effective. 

\begin{table}[t]
    \centering
    \resizebox{0.6\linewidth}{!}{
    \begin{tabular}{c|ccc}
    \hline
    Cross-Attention & MOSE dev & DAVIS 2017 val & SA-V test \\ \hline
    \cref{eq:acrossattn}  & 76.4     & 88.7           & 73.9      \\ \hline
    \cref{eq:ecrossattn}   & 76.5     & 88.6           & 74.0        \\ \hline
    \end{tabular}}
    \caption{\centering Efficient cross-attention variants.}
    \label{tab:cross}
\end{table}
\noindent \textbf{Efficient cross-attention variants.} We compare efficient cross-attention variants. We find that the Linformer variant underperforms the efficient cross-attention in \cref{eq:ecrossattn}, 73.4 vs 74 $\mathcal{J}$\&$\mathcal{F}$ on SA-V test. However, we find that \cref{eq:acrossattn}, can achieve comparable performance, shown in \cref{tab:cross}. 

\begin{table}[t]
    \centering
    \resizebox{0.65\linewidth}{!}{
    \begin{tabular}{c|ccccc}
    \hline
    Resolution & \begin{tabular}[c]{@{}c@{}}MOSE \\ dev\end{tabular} & \begin{tabular}[c]{@{}c@{}}DAVIS \\ 2017 val\end{tabular} & \begin{tabular}[c]{@{}c@{}}SA-V \\ test\end{tabular} & \begin{tabular}[c]{@{}c@{}}FPS \\ A100\end{tabular} & \begin{tabular}[c]{@{}c@{}}Latency (ms) \\ iPhone 15\end{tabular} \\ \hline
    $1024\times 1024$  & 76.5                                                & 89.2                                                      & 74.5                                                 & 85                                                  & 1010.8                                                          \\ \hline
    $512\times 512$    & 74.8                                                & 87.2                                                      & 71.5                                                 & 134                                                 & 80.6                                                            \\ \hline
    \end{tabular}}
    \caption{\centering Ablation study on the effect of input resolution.}
    \label{tab:res}
\end{table}
\noindent \textbf{Impact of input resolution.} We ablate the impact of input resolution for video object segmentation. By default, we used $1024\times 1024$. We experiment with different input resolution, e.g., $512\times 512$. \cref{tab:res} shows that decreasing the input resolution leads to some performance drop. But it improves the efficiency, especially on mobile device, 12.5x speedup on iPhone 15. This gives flexibility for practical deployments with different latency and quality needs.
\vspace{-1mm}
\section{Conclusions}
\vspace{-1mm}
We revisited using a plain, non-hierachical image encoder for building efficient video object segmentation and track anything model, EfficientTAM. With a vanilla lightweight ViT image encoder, EfficientTAM demonstrated competing image and video segmentation capabilities as hierarchical image encoder while being more efficient and deployable on mobile devices. We also proposed an efficient memory module with faster cross-attention, leveraging the locality of spatial memory embeddings. The efficient memory module further improves EfficientTAM's accuracy-efficiency tradeoff on video segmentation and tracking anything. Extensive experiments on semi-supervised video object segmentation, promptable video segmentation, and the segment anything tasks consistently validate the advantages of our EfficientTAM. Our preliminary work suggests that EfficientTAM has many potential applications for on-device tracking anything.
\section{Acknowledgments}
We thank Chaitanya Ryali for valuable discussions and data access support. We thank Ronghang Hu for suggestions. Thanks to Nikhila Ravi for supporting 1 node of A100 for benchmarking. 

\bibliographystyle{assets/plainnat}
\bibliography{main}

\begin{thebibliography}{96}
\providecommand{\natexlab}[1]{#1}
\providecommand{\url}[1]{\texttt{#1}}
\expandafter\ifx\csname urlstyle\endcsname\relax
  \providecommand{\doi}[1]{doi: #1}\else
  \providecommand{\doi}{doi: \begingroup \urlstyle{rm}\Url}\fi

\bibitem[Beltagy et~al.(2020)Beltagy, Peters, and Cohan]{beltagy2020longformer}
Iz~Beltagy, Matthew~E. Peters, and Arman Cohan.
\newblock Longformer: The long-document transformer.
\newblock \emph{arXiv:2004.05150}, 2020.

\bibitem[Bhat et~al.(2020)Bhat, Lawin, Danelljan, Robinson, Felsberg, Van~Gool, and Timofte]{bhat2020learning}
Goutam Bhat, Felix~J{\"a}remo Lawin, Martin Danelljan, Andreas Robinson, Michael Felsberg, Luc Van~Gool, and Radu Timofte.
\newblock Learning what to learn for video object segmentation.
\newblock In \emph{Computer Vision--ECCV 2020: 16th European Conference, Glasgow, UK, August 23--28, 2020, Proceedings, Part II 16}, pages 777--794. Springer, 2020.

\bibitem[Brox and Malik(2010)]{brox2010object}
Thomas Brox and Jitendra Malik.
\newblock Object segmentation by long term analysis of point trajectories.
\newblock In \emph{European conference on computer vision}, pages 282--295. Springer, 2010.

\bibitem[Caelles et~al.(2018)Caelles, Montes, Maninis, Chen, Van~Gool, Perazzi, and Pont-Tuset]{caelles20182018}
Sergi Caelles, Alberto Montes, Kevis-Kokitsi Maninis, Yuhua Chen, Luc Van~Gool, Federico Perazzi, and Jordi Pont-Tuset.
\newblock The 2018 davis challenge on video object segmentation.
\newblock \emph{arXiv preprint arXiv:1803.00557}, 2018.

\bibitem[Cai et~al.(2023)Cai, Li, Hu, Gan, and Han]{cai2023efficientvit}
Han Cai, Junyan Li, Muyan Hu, Chuang Gan, and Song Han.
\newblock Efficientvit: Lightweight multi-scale attention for high-resolution dense prediction.
\newblock In \emph{Proceedings of the IEEE/CVF International Conference on Computer Vision}, pages 17302--17313, 2023.

\bibitem[Cen et~al.(2023)Cen, Wu, Wang, Li, Yang, Pei, Kong, Liu, and Chen]{cen2023sad}
Jun Cen, Yizheng Wu, Kewei Wang, Xingyi Li, Jingkang Yang, Yixuan Pei, Lingdong Kong, Ziwei Liu, and Qifeng Chen.
\newblock Sad: Segment any rgbd.
\newblock \emph{arXiv preprint arXiv:2305.14207}, 2023.

\bibitem[Chen et~al.(2023{\natexlab{a}})Chen, Yang, and Zhang]{chen2023semantic}
Jiaqi Chen, Zeyu Yang, and Li~Zhang.
\newblock Semantic segment anything.
\newblock \url{https://github.com/fudan-zvg/Semantic-Segment-Anything}, 2023{\natexlab{a}}.

\bibitem[Chen et~al.(2023{\natexlab{b}})Chen, Zhu, Deng, Cao, Wang, Zhang, Li, Sun, Zang, and Mao]{chen2023sam}
Tianrun Chen, Lanyun Zhu, Chaotao Deng, Runlong Cao, Yan Wang, Shangzhan Zhang, Zejian Li, Lingyun Sun, Ying Zang, and Papa Mao.
\newblock Sam-adapter: Adapting segment anything in underperformed scenes.
\newblock In \emph{Proceedings of the IEEE/CVF International Conference on Computer Vision}, pages 3367--3375, 2023{\natexlab{b}}.

\bibitem[Cheng et~al.(2022)Cheng, Misra, Schwing, Kirillov, and Girdhar]{cheng2022masked}
Bowen Cheng, Ishan Misra, Alexander~G Schwing, Alexander Kirillov, and Rohit Girdhar.
\newblock Masked-attention mask transformer for universal image segmentation.
\newblock In \emph{Proceedings of the IEEE/CVF conference on computer vision and pattern recognition}, pages 1290--1299, 2022.

\bibitem[Cheng and Schwing(2022)]{cheng2022xmem}
Ho~Kei Cheng and Alexander~G Schwing.
\newblock Xmem: Long-term video object segmentation with an atkinson-shiffrin memory model.
\newblock In \emph{European Conference on Computer Vision}, pages 640--658. Springer, 2022.

\bibitem[Cheng et~al.(2021{\natexlab{a}})Cheng, Tai, and Tang]{cheng2021modular}
Ho~Kei Cheng, Yu-Wing Tai, and Chi-Keung Tang.
\newblock Modular interactive video object segmentation: Interaction-to-mask, propagation and difference-aware fusion.
\newblock In \emph{Proceedings of the IEEE/CVF Conference on Computer Vision and Pattern Recognition}, pages 5559--5568, 2021{\natexlab{a}}.

\bibitem[Cheng et~al.(2021{\natexlab{b}})Cheng, Tai, and Tang]{cheng2021rethinking}
Ho~Kei Cheng, Yu-Wing Tai, and Chi-Keung Tang.
\newblock Rethinking space-time networks with improved memory coverage for efficient video object segmentation.
\newblock \emph{Advances in Neural Information Processing Systems}, 34:\penalty0 11781--11794, 2021{\natexlab{b}}.

\bibitem[Cheng et~al.(2023{\natexlab{a}})Cheng, Oh, Price, Schwing, and Lee]{cheng2023tracking}
Ho~Kei Cheng, Seoung~Wug Oh, Brian Price, Alexander Schwing, and Joon-Young Lee.
\newblock Tracking anything with decoupled video segmentation.
\newblock In \emph{Proceedings of the IEEE/CVF International Conference on Computer Vision}, pages 1316--1326, 2023{\natexlab{a}}.

\bibitem[Cheng et~al.(2024)Cheng, Oh, Price, Lee, and Schwing]{cheng2024putting}
Ho~Kei Cheng, Seoung~Wug Oh, Brian Price, Joon-Young Lee, and Alexander Schwing.
\newblock Putting the object back into video object segmentation.
\newblock In \emph{Proceedings of the IEEE/CVF Conference on Computer Vision and Pattern Recognition}, pages 3151--3161, 2024.

\bibitem[Cheng et~al.(2023{\natexlab{b}})Cheng, Li, Xu, Li, Yang, Wang, and Yang]{cheng2023segment}
Yangming Cheng, Liulei Li, Yuanyou Xu, Xiaodi Li, Zongxin Yang, Wenguan Wang, and Yi~Yang.
\newblock Segment and track anything.
\newblock \emph{arXiv preprint arXiv:2305.06558}, 2023{\natexlab{b}}.

\bibitem[Choromanski et~al.(2020)Choromanski, Likhosherstov, Dohan, Song, Gane, Sarlos, Hawkins, Davis, Mohiuddin, Kaiser, et~al.]{choromanski2020rethinking}
Krzysztof Choromanski, Valerii Likhosherstov, David Dohan, Xingyou Song, Andreea Gane, Tamas Sarlos, Peter Hawkins, Jared Davis, Afroz Mohiuddin, Lukasz Kaiser, et~al.
\newblock Rethinking attention with performers.
\newblock \emph{arXiv preprint arXiv:2009.14794}, 2020.

\bibitem[Clark et~al.(2020)Clark, Luong, Le, and Manning]{clark2020electra}
Kevin Clark, Minh-Thang Luong, Quoc~V. Le, and Christopher~D. Manning.
\newblock {ELECTRA}: Pre-training text encoders as discriminators rather than generators.
\newblock In \emph{ICLR}, 2020.
\newblock \url{https://openreview.net/pdf?id=r1xMH1BtvB}.

\bibitem[Delatolas et~al.(2024)Delatolas, Kalogeiton, and Papadopoulos]{delatolas2024learning}
Thanos Delatolas, Vicky Kalogeiton, and Dim~P Papadopoulos.
\newblock Learning the what and how of annotation in video object segmentation.
\newblock In \emph{Proceedings of the IEEE/CVF Winter Conference on Applications of Computer Vision}, pages 6951--6961, 2024.

\bibitem[Deng et~al.(2023)Deng, Cui, Liu, Yao, Remedios, Bao, Landman, Wheless, Coburn, Wilson, et~al.]{deng2023segment}
Ruining Deng, Can Cui, Quan Liu, Tianyuan Yao, Lucas~W Remedios, Shunxing Bao, Bennett~A Landman, Lee~E Wheless, Lori~A Coburn, Keith~T Wilson, et~al.
\newblock Segment anything model (sam) for digital pathology: Assess zero-shot segmentation on whole slide imaging.
\newblock \emph{arXiv preprint arXiv:2304.04155}, 2023.

\bibitem[Ding et~al.(2023)Ding, Liu, He, Jiang, Torr, and Bai]{ding2023mose}
Henghui Ding, Chang Liu, Shuting He, Xudong Jiang, Philip~HS Torr, and Song Bai.
\newblock Mose: A new dataset for video object segmentation in complex scenes.
\newblock In \emph{Proceedings of the IEEE/CVF International Conference on Computer Vision}, pages 20224--20234, 2023.

\bibitem[Ding et~al.(2024)Ding, Qian, Dong, Zhang, Zang, Cao, Guo, Lin, and Wang]{ding2024sam2long}
Shuangrui Ding, Rui Qian, Xiaoyi Dong, Pan Zhang, Yuhang Zang, Yuhang Cao, Yuwei Guo, Dahua Lin, and Jiaqi Wang.
\newblock Sam2long: Enhancing sam 2 for long video segmentation with a training-free memory tree.
\newblock \emph{arXiv preprint arXiv:2410.16268}, 2024.

\bibitem[Dosovitskiy et~al.(2020)Dosovitskiy, Beyer, Kolesnikov, Weissenborn, Zhai, Unterthiner, Dehghani, Minderer, Heigold, Gelly, et~al.]{dosovitskiy2020image}
Alexey Dosovitskiy, Lucas Beyer, Alexander Kolesnikov, Dirk Weissenborn, Xiaohua Zhai, Thomas Unterthiner, Mostafa Dehghani, Matthias Minderer, Georg Heigold, Sylvain Gelly, et~al.
\newblock An image is worth 16x16 words: Transformers for image recognition at scale.
\newblock In \emph{International Conference on Learning Representations}, 2020.

\bibitem[Duke et~al.(2021)Duke, Ahmed, Wolf, Aarabi, and Taylor]{duke2021sstvos}
Brendan Duke, Abdalla Ahmed, Christian Wolf, Parham Aarabi, and Graham~W Taylor.
\newblock Sstvos: Sparse spatiotemporal transformers for video object segmentation.
\newblock In \emph{Proceedings of the IEEE/CVF conference on computer vision and pattern recognition}, pages 5912--5921, 2021.

\bibitem[Faktor and Irani(2014)]{faktor2014video}
Alon Faktor and Michal Irani.
\newblock Video segmentation by non-local consensus voting.
\newblock In \emph{BMVC}, volume~2, page~8, 2014.

\bibitem[Fan et~al.(2021)Fan, Xiong, Mangalam, Li, Yan, Malik, and Feichtenhofer]{fan2021multiscale}
Haoqi Fan, Bo~Xiong, Karttikeya Mangalam, Yanghao Li, Zhicheng Yan, Jitendra Malik, and Christoph Feichtenhofer.
\newblock Multiscale vision transformers.
\newblock In \emph{Proceedings of the IEEE/CVF international conference on computer vision}, pages 6824--6835, 2021.

\bibitem[Fragkiadaki et~al.(2012)Fragkiadaki, Zhang, and Shi]{fragkiadaki2012video}
Katerina Fragkiadaki, Geng Zhang, and Jianbo Shi.
\newblock Video segmentation by tracing discontinuities in a trajectory embedding.
\newblock In \emph{2012 IEEE Conference on Computer Vision and Pattern Recognition}, pages 1846--1853. IEEE, 2012.

\bibitem[Gao et~al.(2023)Gao, Lin, Xie, Zhou, Cheng, and Yan]{gao2023editanything}
Shanghua Gao, Zhijie Lin, Xingyu Xie, Pan Zhou, Ming-Ming Cheng, and Shuicheng Yan.
\newblock Editanything: Empowering unparalleled flexibility in image editing and generation.
\newblock In \emph{Proceedings of the 31st ACM International Conference on Multimedia}, pages 9414--9416, 2023.

\bibitem[Graham et~al.(2021)Graham, El-Nouby, Touvron, Stock, Joulin, J{\'e}gou, and Douze]{graham2021levit}
Benjamin Graham, Alaaeldin El-Nouby, Hugo Touvron, Pierre Stock, Armand Joulin, Herv{\'e} J{\'e}gou, and Matthijs Douze.
\newblock Levit: a vision transformer in convnet's clothing for faster inference.
\newblock In \emph{Proceedings of the IEEE/CVF international conference on computer vision}, pages 12259--12269, 2021.

\bibitem[Grundmann et~al.(2010)Grundmann, Kwatra, Han, and Essa]{grundmann2010efficient}
Matthias Grundmann, Vivek Kwatra, Mei Han, and Irfan Essa.
\newblock Efficient hierarchical graph-based video segmentation.
\newblock In \emph{2010 ieee computer society conference on computer vision and pattern recognition}, pages 2141--2148. IEEE, 2010.

\bibitem[Han et~al.(2023)Han, Zhang, Qiao, Qamar, Jung, Lee, Bae, and Hong]{han2023segment}
Dongsheng Han, Chaoning Zhang, Yu~Qiao, Maryam Qamar, Yuna Jung, SeungKyu Lee, Sung-Ho Bae, and Choong~Seon Hong.
\newblock Segment anything model (sam) meets glass: Mirror and transparent objects cannot be easily detected.
\newblock \emph{arXiv preprint arXiv:2305.00278}, 2023.

\bibitem[Heo et~al.(2020)Heo, Jun~Koh, and Kim]{heo2020interactive}
Yuk Heo, Yeong Jun~Koh, and Chang-Su Kim.
\newblock Interactive video object segmentation using global and local transfer modules.
\newblock In \emph{Computer Vision--ECCV 2020: 16th European Conference, Glasgow, UK, August 23--28, 2020, Proceedings, Part XVII 16}, pages 297--313. Springer, 2020.

\bibitem[Homayounfar et~al.(2021)Homayounfar, Liang, Ma, and Urtasun]{homayounfar2021videoclick}
Namdar Homayounfar, Justin Liang, Wei-Chiu Ma, and Raquel Urtasun.
\newblock Videoclick: Video object segmentation with a single click.
\newblock \emph{arXiv preprint arXiv:2101.06545}, 2021.

\bibitem[Hong et~al.(2024)Hong, Liu, Chen, Tan, Feng, Zhou, Guo, Li, Chen, Gao, et~al.]{hong2024lvos}
Lingyi Hong, Zhongying Liu, Wenchao Chen, Chenzhi Tan, Yuang Feng, Xinyu Zhou, Pinxue Guo, Jinglun Li, Zhaoyu Chen, Shuyong Gao, et~al.
\newblock Lvos: A benchmark for large-scale long-term video object segmentation.
\newblock \emph{arXiv preprint arXiv:2404.19326}, 2024.

\bibitem[Jiang and Holz(2023)]{jiang2023restore}
Jiaxi Jiang and Christian Holz.
\newblock Restore anything pipeline: Segment anything meets image restoration.
\newblock \emph{arXiv preprint arXiv:2305.13093}, 2023.

\bibitem[Katharopoulos et~al.(2020)Katharopoulos, Vyas, Pappas, and Fleuret]{katharopoulos-et-al-2020}
A.~Katharopoulos, A.~Vyas, N.~Pappas, and F.~Fleuret.
\newblock Transformers are rnns: Fast autoregressive transformers with linear attention.
\newblock In \emph{Proceedings of the International Conference on Machine Learning (ICML)}, 2020.

\bibitem[Kirillov et~al.(2023)Kirillov, Mintun, Ravi, Mao, Rolland, Gustafson, Xiao, Whitehead, Berg, Lo, et~al.]{kirillov2023segment}
Alexander Kirillov, Eric Mintun, Nikhila Ravi, Hanzi Mao, Chloe Rolland, Laura Gustafson, Tete Xiao, Spencer Whitehead, Alexander~C Berg, Wan-Yen Lo, et~al.
\newblock Segment anything.
\newblock In \emph{Proceedings of the IEEE/CVF International Conference on Computer Vision}, pages 4015--4026, 2023.

\bibitem[LeCun et~al.(1989)LeCun, Boser, Denker, Henderson, Howard, Hubbard, and Jackel]{lecun1989backpropagation}
Yann LeCun, Bernhard Boser, John~S Denker, Donnie Henderson, Richard~E Howard, Wayne Hubbard, and Lawrence~D Jackel.
\newblock Backpropagation applied to handwritten zip code recognition.
\newblock \emph{Neural computation}, 1\penalty0 (4):\penalty0 541--551, 1989.

\bibitem[Lee et~al.(2011)Lee, Kim, and Grauman]{lee2011key}
Yong~Jae Lee, Jaechul Kim, and Kristen Grauman.
\newblock Key-segments for video object segmentation.
\newblock In \emph{2011 International conference on computer vision}, pages 1995--2002. IEEE, 2011.

\bibitem[Li et~al.(2013)Li, Kim, Humayun, Tsai, and Rehg]{li2013video}
Fuxin Li, Taeyoung Kim, Ahmad Humayun, David Tsai, and James~M Rehg.
\newblock Video segmentation by tracking many figure-ground segments.
\newblock In \emph{Proceedings of the IEEE international conference on computer vision}, pages 2192--2199, 2013.

\bibitem[Li et~al.(2022{\natexlab{a}})Li, Xia, Li, Li, Wang, Xiao, Wang, Zheng, and Pan]{li2022next}
Jiashi Li, Xin Xia, Wei Li, Huixia Li, Xing Wang, Xuefeng Xiao, Rui Wang, Min Zheng, and Xin Pan.
\newblock Next-vit: Next generation vision transformer for efficient deployment in realistic industrial scenarios.
\newblock \emph{arXiv preprint arXiv:2207.05501}, 2022{\natexlab{a}}.

\bibitem[Li et~al.(2022{\natexlab{b}})Li, Hu, Xiong, Zhang, Pan, and Liu]{li2022recurrent}
Mingxing Li, Li~Hu, Zhiwei Xiong, Bang Zhang, Pan Pan, and Dong Liu.
\newblock Recurrent dynamic embedding for video object segmentation.
\newblock In \emph{Proceedings of the IEEE/CVF Conference on Computer Vision and Pattern Recognition}, pages 1332--1341, 2022{\natexlab{b}}.

\bibitem[Li et~al.(2022{\natexlab{c}})Li, Mao, Girshick, and He]{li2022exploring}
Yanghao Li, Hanzi Mao, Ross Girshick, and Kaiming He.
\newblock Exploring plain vision transformer backbones for object detection.
\newblock In \emph{European conference on computer vision}, pages 280--296. Springer, 2022{\natexlab{c}}.

\bibitem[Li et~al.(2022{\natexlab{d}})Li, Wu, Fan, Mangalam, Xiong, Malik, and Feichtenhofer]{li2022mvitv2}
Yanghao Li, Chao-Yuan Wu, Haoqi Fan, Karttikeya Mangalam, Bo~Xiong, Jitendra Malik, and Christoph Feichtenhofer.
\newblock Mvitv2: Improved multiscale vision transformers for classification and detection.
\newblock In \emph{Proceedings of the IEEE/CVF conference on computer vision and pattern recognition}, pages 4804--4814, 2022{\natexlab{d}}.

\bibitem[Li et~al.(2022{\natexlab{e}})Li, Yuan, Wen, Hu, Evangelidis, Tulyakov, Wang, and Ren]{li2022efficientformer}
Yanyu Li, Geng Yuan, Yang Wen, Ju~Hu, Georgios Evangelidis, Sergey Tulyakov, Yanzhi Wang, and Jian Ren.
\newblock Efficientformer: Vision transformers at mobilenet speed.
\newblock \emph{Advances in Neural Information Processing Systems}, 35:\penalty0 12934--12949, 2022{\natexlab{e}}.

\bibitem[Liu et~al.(2023{\natexlab{a}})Liu, Zeng, Ren, Li, Zhang, Yang, Jiang, Li, Yang, Su, et~al.]{liu2023grounding}
Shilong Liu, Zhaoyang Zeng, Tianhe Ren, Feng Li, Hao Zhang, Jie Yang, Qing Jiang, Chunyuan Li, Jianwei Yang, Hang Su, et~al.
\newblock Grounding dino: Marrying dino with grounded pre-training for open-set object detection.
\newblock \emph{arXiv preprint arXiv:2303.05499}, 2023{\natexlab{a}}.

\bibitem[Liu et~al.(2023{\natexlab{b}})Liu, Peng, Zheng, Yang, Hu, and Yuan]{liu2023efficientvit}
Xinyu Liu, Houwen Peng, Ningxin Zheng, Yuqing Yang, Han Hu, and Yixuan Yuan.
\newblock Efficientvit: Memory efficient vision transformer with cascaded group attention.
\newblock In \emph{Proceedings of the IEEE/CVF Conference on Computer Vision and Pattern Recognition}, pages 14420--14430, 2023{\natexlab{b}}.

\bibitem[Liu et~al.(2021)Liu, Lin, Cao, Hu, Wei, Zhang, Lin, and Guo]{liu2021swin}
Ze~Liu, Yutong Lin, Yue Cao, Han Hu, Yixuan Wei, Zheng Zhang, Stephen Lin, and Baining Guo.
\newblock Swin transformer: Hierarchical vision transformer using shifted windows.
\newblock In \emph{Proceedings of the IEEE/CVF international conference on computer vision}, pages 10012--10022, 2021.

\bibitem[Loshchilov and Hutter(2019)]{loshchilov2017decoupled}
Ilya Loshchilov and Frank Hutter.
\newblock Decoupled weight decay regularization.
\newblock 2019.

\bibitem[Ma and Wang(2023)]{ma2023segment}
Jun Ma and Bo~Wang.
\newblock Segment anything in medical images.
\newblock \emph{arXiv preprint arXiv:2304.12306}, 2023.

\bibitem[Mehta and Rastegari(2021)]{mehta2021mobilevit}
Sachin Mehta and Mohammad Rastegari.
\newblock Mobilevit: Light-weight, general-purpose, and mobile-friendly vision transformer.
\newblock In \emph{International Conference on Learning Representations}, 2021.

\bibitem[Oh et~al.(2019)Oh, Lee, Xu, and Kim]{oh2019video}
Seoung~Wug Oh, Joon-Young Lee, Ning Xu, and Seon~Joo Kim.
\newblock Video object segmentation using space-time memory networks.
\newblock In \emph{Proceedings of the IEEE/CVF international conference on computer vision}, pages 9226--9235, 2019.

\bibitem[Papazoglou and Ferrari(2013)]{papazoglou2013fast}
Anestis Papazoglou and Vittorio Ferrari.
\newblock Fast object segmentation in unconstrained video.
\newblock In \emph{Proceedings of the IEEE international conference on computer vision}, pages 1777--1784, 2013.

\bibitem[Perazzi et~al.(2012)Perazzi, Kr{\"a}henb{\"u}hl, Pritch, and Hornung]{perazzi2012saliency}
Federico Perazzi, Philipp Kr{\"a}henb{\"u}hl, Yael Pritch, and Alexander Hornung.
\newblock Saliency filters: Contrast based filtering for salient region detection.
\newblock In \emph{2012 IEEE conference on computer vision and pattern recognition}, pages 733--740. IEEE, 2012.

\bibitem[Perazzi et~al.(2016)Perazzi, Pont-Tuset, McWilliams, Van~Gool, Gross, and Sorkine-Hornung]{perazzi2016benchmark}
Federico Perazzi, Jordi Pont-Tuset, Brian McWilliams, Luc Van~Gool, Markus Gross, and Alexander Sorkine-Hornung.
\newblock A benchmark dataset and evaluation methodology for video object segmentation.
\newblock In \emph{Proceedings of the IEEE conference on computer vision and pattern recognition}, pages 724--732, 2016.

\bibitem[Pont-Tuset et~al.(2017)Pont-Tuset, Perazzi, Caelles, Arbel{\'a}ez, Sorkine-Hornung, and Van~Gool]{pont20172017}
Jordi Pont-Tuset, Federico Perazzi, Sergi Caelles, Pablo Arbel{\'a}ez, Alex Sorkine-Hornung, and Luc Van~Gool.
\newblock The 2017 davis challenge on video object segmentation.
\newblock \emph{arXiv preprint arXiv:1704.00675}, 2017.

\bibitem[Qin et~al.(2024)Qin, Leichner, Delakis, Fornoni, Luo, Yang, Wang, Banbury, Ye, Akin, et~al.]{qin2024mobilenetv4}
Danfeng Qin, Chas Leichner, Manolis Delakis, Marco Fornoni, Shixin Luo, Fan Yang, Weijun Wang, Colby Banbury, Chengxi Ye, Berkin Akin, et~al.
\newblock Mobilenetv4-universal models for the mobile ecosystem.
\newblock \emph{arXiv preprint arXiv:2404.10518}, 2024.

\bibitem[Qiu et~al.(2024)Qiu, Liu, Li, Zhang, and Li]{qiu2024ded}
Junlong Qiu, Wei Liu, Erzhu Li, Lianpeng Zhang, and Xing Li.
\newblock Ded-sam: Adapting segment anything model 2 for dual encoder-decoder change detection.
\newblock \emph{IEEE Journal of Selected Topics in Applied Earth Observations and Remote Sensing}, 2024.

\bibitem[Raji{\v{c}} et~al.(2023)Raji{\v{c}}, Ke, Tai, Tang, Danelljan, and Yu]{rajivc2023segment}
Frano Raji{\v{c}}, Lei Ke, Yu-Wing Tai, Chi-Keung Tang, Martin Danelljan, and Fisher Yu.
\newblock Segment anything meets point tracking.
\newblock \emph{arXiv preprint arXiv:2307.01197}, 2023.

\bibitem[Ravi et~al.(2024)Ravi, Gabeur, Hu, Hu, Ryali, Ma, Khedr, R{\"a}dle, Rolland, Gustafson, et~al.]{ravi2024sam}
Nikhila Ravi, Valentin Gabeur, Yuan-Ting Hu, Ronghang Hu, Chaitanya Ryali, Tengyu Ma, Haitham Khedr, Roman R{\"a}dle, Chloe Rolland, Laura Gustafson, et~al.
\newblock Sam 2: Segment anything in images and videos.
\newblock \emph{arXiv preprint arXiv:2408.00714}, 2024.

\bibitem[Robinson et~al.(2020)Robinson, Lawin, Danelljan, Khan, and Felsberg]{robinson2020learning}
Andreas Robinson, Felix~Jaremo Lawin, Martin Danelljan, Fahad~Shahbaz Khan, and Michael Felsberg.
\newblock Learning fast and robust target models for video object segmentation.
\newblock In \emph{Proceedings of the IEEE/CVF conference on computer vision and pattern recognition}, pages 7406--7415, 2020.

\bibitem[Ryali et~al.(2023)Ryali, Hu, Bolya, Wei, Fan, Huang, Aggarwal, Chowdhury, Poursaeed, Hoffman, et~al.]{ryali2023hiera}
Chaitanya Ryali, Yuan-Ting Hu, Daniel Bolya, Chen Wei, Haoqi Fan, Po-Yao Huang, Vaibhav Aggarwal, Arkabandhu Chowdhury, Omid Poursaeed, Judy Hoffman, et~al.
\newblock Hiera: A hierarchical vision transformer without the bells-and-whistles.
\newblock In \emph{International Conference on Machine Learning}, pages 29441--29454. PMLR, 2023.

\bibitem[Shen et~al.(2024)Shen, Ding, Shao, and Unberath]{shen2024performance}
Yiqing Shen, Hao Ding, Xinyuan Shao, and Mathias Unberath.
\newblock Performance and non-adversarial robustness of the segment anything model 2 in surgical video segmentation.
\newblock \emph{arXiv preprint arXiv:2408.04098}, 2024.

\bibitem[Shen et~al.(2018)Shen, Zhang, Zhao, Yi, and Li]{shen2018efficient}
Zhuoran Shen, Mingyuan Zhang, Haiyu Zhao, Shuai Yi, and Hongsheng Li.
\newblock Efficient attention: Attention with linear complexities.
\newblock \emph{arXiv preprint arXiv:1812.01243}, 2018.

\bibitem[Sun et~al.(2023)Sun, Ma, Yuan, and Wang]{sun2023explain}
Ao~Sun, Pingchuan Ma, Yuanyuan Yuan, and Shuai Wang.
\newblock Explain any concept: Segment anything meets concept-based explanation.
\newblock \emph{arXiv preprint arXiv:2305.10289}, 2023.

\bibitem[Tang et~al.(2024)Tang, Zhao, Ford, Benhaim, and Zhang]{tang2024segment}
George Tang, William Zhao, Logan Ford, David Benhaim, and Paul Zhang.
\newblock Segment any mesh: Zero-shot mesh part segmentation via lifting segment anything 2 to 3d.
\newblock \emph{arXiv preprint arXiv:2408.13679}, 2024.

\bibitem[Tang et~al.(2023)Tang, Xiao, and Li]{tang2023can}
Lv~Tang, Haoke Xiao, and Bo~Li.
\newblock Can sam segment anything? when sam meets camouflaged object detection.
\newblock \emph{arXiv preprint arXiv:2304.04709}, 2023.

\bibitem[Tariq et~al.(2023)Tariq, Arfeto, Zhang, and Shin]{tariq2023segment}
Shehbaz Tariq, Brian~Estadimas Arfeto, Chaoning Zhang, and Hyundong Shin.
\newblock Segment anything meets semantic communication.
\newblock \emph{arXiv preprint arXiv:2306.02094}, 2023.

\bibitem[Taylor et~al.(2015)Taylor, Karasev, and Soatto]{taylor2015causal}
Brian Taylor, Vasiliy Karasev, and Stefano Soatto.
\newblock Causal video object segmentation from persistence of occlusions.
\newblock In \emph{Proceedings of the IEEE conference on computer vision and pattern recognition}, pages 4268--4276, 2015.

\bibitem[Touvron et~al.(2021)Touvron, Cord, Douze, Massa, Sablayrolles, and J{\'e}gou]{touvron2021training}
Hugo Touvron, Matthieu Cord, Matthijs Douze, Francisco Massa, Alexandre Sablayrolles, and Herv{\'e} J{\'e}gou.
\newblock Training data-efficient image transformers \& distillation through attention.
\newblock In \emph{International conference on machine learning}, pages 10347--10357. PMLR, 2021.

\bibitem[Vaswani et~al.(2017)Vaswani, Shazeer, Parmar, Uszkoreit, Jones, Gomez, Kaiser, and Polosukhin]{attention_is_all_you_need}
Ashish Vaswani, Noam Shazeer, Niki Parmar, Jakob Uszkoreit, Llion Jones, Aidan~N. Gomez, \L{}ukasz Kaiser, and Illia Polosukhin.
\newblock Attention is all you need.
\newblock In \emph{Proceedings of the 31st International Conference on Neural Information Processing Systems}, NIPS'17, page 6000–6010, Red Hook, NY, USA, 2017. Curran Associates Inc.
\newblock ISBN 9781510860964.

\bibitem[Wang et~al.(2023)Wang, Chen, Wu, Luo, Tang, Dai, Zhao, Xie, Yuan, and Jiang]{wang2023look}
Junke Wang, Dongdong Chen, Zuxuan Wu, Chong Luo, Chuanxin Tang, Xiyang Dai, Yucheng Zhao, Yujia Xie, Lu~Yuan, and Yu-Gang Jiang.
\newblock Look before you match: Instance understanding matters in video object segmentation.
\newblock In \emph{Proceedings of the IEEE/CVF conference on computer vision and pattern recognition}, pages 2268--2278, 2023.

\bibitem[Wang et~al.(2020)Wang, Li, Khabsa, Fang, and Ma]{wang2020linformer}
Sinong Wang, Belinda Li, Madian Khabsa, Han Fang, and Hao Ma.
\newblock Linformer: Self-attention with linear complexity.
\newblock \emph{arXiv preprint arXiv:2006.04768}, 2020.

\bibitem[Wang et~al.(2015)Wang, Shen, and Porikli]{wang2015saliency}
Wenguan Wang, Jianbing Shen, and Fatih Porikli.
\newblock Saliency-aware geodesic video object segmentation.
\newblock In \emph{Proceedings of the IEEE conference on computer vision and pattern recognition}, pages 3395--3402, 2015.

\bibitem[Wang et~al.(2021)Wang, Xie, Li, Fan, Song, Liang, Lu, Luo, and Shao]{wang2021pyramid}
Wenhai Wang, Enze Xie, Xiang Li, Deng-Ping Fan, Kaitao Song, Ding Liang, Tong Lu, Ping Luo, and Ling Shao.
\newblock Pyramid vision transformer: A versatile backbone for dense prediction without convolutions.
\newblock In \emph{Proceedings of the IEEE/CVF international conference on computer vision}, pages 568--578, 2021.

\bibitem[Wu et~al.(2022)Wu, Zhang, Peng, Liu, Xiao, Fu, and Yuan]{wu2022tinyvit}
Kan Wu, Jinnian Zhang, Houwen Peng, Mengchen Liu, Bin Xiao, Jianlong Fu, and Lu~Yuan.
\newblock Tinyvit: Fast pretraining distillation for small vision transformers.
\newblock In \emph{European Conference on Computer Vision}, pages 68--85. Springer, 2022.

\bibitem[Wu et~al.(2023)Wu, Yang, Wu, and Chan]{wu2023scalable}
Qiangqiang Wu, Tianyu Yang, Wei Wu, and Antoni~B Chan.
\newblock Scalable video object segmentation with simplified framework.
\newblock In \emph{Proceedings of the IEEE/CVF International Conference on Computer Vision}, pages 13879--13889, 2023.

\bibitem[Xiong et~al.(2024{\natexlab{a}})Xiong, Wu, Tan, Li, Tang, Chen, Li, Ma, and Li]{xiong2024sam2}
Xinyu Xiong, Zihuang Wu, Shuangyi Tan, Wenxue Li, Feilong Tang, Ying Chen, Siying Li, Jie Ma, and Guanbin Li.
\newblock Sam2-unet: Segment anything 2 makes strong encoder for natural and medical image segmentation.
\newblock \emph{arXiv preprint arXiv:2408.08870}, 2024{\natexlab{a}}.

\bibitem[Xiong et~al.(2021)Xiong, Zeng, Chakraborty, Tan, Fung, Li, and Singh]{xiong2021nystromformer}
Yunyang Xiong, Zhanpeng Zeng, Rudrasis Chakraborty, Mingxing Tan, Glenn Fung, Yin Li, and Vikas Singh.
\newblock Nystr{\"o}mformer: A nystr{\"o}m-based algorithm for approximating self-attention.
\newblock In \emph{Proceedings of the AAAI Conference on Artificial Intelligence}, volume~35, pages 14138--14148, 2021.

\bibitem[Xiong et~al.(2024{\natexlab{b}})Xiong, Varadarajan, Wu, Xiang, Xiao, Zhu, Dai, Wang, Sun, Iandola, et~al.]{xiong2024efficientsam}
Yunyang Xiong, Bala Varadarajan, Lemeng Wu, Xiaoyu Xiang, Fanyi Xiao, Chenchen Zhu, Xiaoliang Dai, Dilin Wang, Fei Sun, Forrest Iandola, et~al.
\newblock Efficientsam: Leveraged masked image pretraining for efficient segment anything.
\newblock In \emph{Proceedings of the IEEE/CVF Conference on Computer Vision and Pattern Recognition}, pages 16111--16121, 2024{\natexlab{b}}.

\bibitem[Xu and Corso(2012)]{xu2012evaluation}
Chenliang Xu and Jason~J Corso.
\newblock Evaluation of super-voxel methods for early video processing.
\newblock In \emph{2012 IEEE conference on computer vision and pattern recognition}, pages 1202--1209. IEEE, 2012.

\bibitem[Xu et~al.(2018)Xu, Yang, Fan, Yang, Yue, Liang, Price, Cohen, and Huang]{xu2018youtube}
Ning Xu, Linjie Yang, Yuchen Fan, Jianchao Yang, Dingcheng Yue, Yuchen Liang, Brian Price, Scott Cohen, and Thomas Huang.
\newblock Youtube-vos: Sequence-to-sequence video object segmentation.
\newblock In \emph{Proceedings of the European conference on computer vision (ECCV)}, pages 585--601, 2018.

\bibitem[Yang et~al.(2023)Yang, Gao, Li, Gao, Wang, and Zheng]{yang2023track}
Jinyu Yang, Mingqi Gao, Zhe Li, Shang Gao, Fangjing Wang, and Feng Zheng.
\newblock Track anything: Segment anything meets videos.
\newblock \emph{arXiv preprint arXiv:2304.11968}, 2023.

\bibitem[Yang and Yang(2022)]{yang2022decoupling}
Zongxin Yang and Yi~Yang.
\newblock Decoupling features in hierarchical propagation for video object segmentation.
\newblock \emph{Advances in Neural Information Processing Systems}, 35:\penalty0 36324--36336, 2022.

\bibitem[Yang et~al.(2024)Yang, Miao, Wei, Wang, Wang, and Yang]{yang2024scalable}
Zongxin Yang, Jiaxu Miao, Yunchao Wei, Wenguan Wang, Xiaohan Wang, and Yi~Yang.
\newblock Scalable video object segmentation with identification mechanism.
\newblock \emph{IEEE Transactions on Pattern Analysis and Machine Intelligence}, 2024.

\bibitem[You et~al.(2023)You, Xiong, Dai, Wu, Zhang, Fan, Vajda, and Lin]{you2023castling}
Haoran You, Yunyang Xiong, Xiaoliang Dai, Bichen Wu, Peizhao Zhang, Haoqi Fan, Peter Vajda, and Yingyan~Celine Lin.
\newblock Castling-vit: Compressing self-attention via switching towards linear-angular attention at vision transformer inference.
\newblock In \emph{Proceedings of the IEEE/CVF Conference on Computer Vision and Pattern Recognition}, pages 14431--14442, 2023.

\bibitem[Yu et~al.(2023)Yu, Feng, Feng, Liu, Jin, Zeng, and Chen]{yu2023inpaint}
Tao Yu, Runseng Feng, Ruoyu Feng, Jinming Liu, Xin Jin, Wenjun Zeng, and Zhibo Chen.
\newblock Inpaint anything: Segment anything meets image inpainting.
\newblock \emph{arXiv preprint arXiv:2304.06790}, 2023.

\bibitem[Zaheer et~al.(2020)Zaheer, Guruganesh, Dubey, Ainslie, Alberti, Ontanon, Pham, Ravula, Wang, Yang, et~al.]{zaheer2020bigbird}
Manzil Zaheer, Guru Guruganesh, Avinava Dubey, Joshua Ainslie, Chris Alberti, Santiago Ontanon, Philip Pham, Anirudh Ravula, Qifan Wang, Li~Yang, et~al.
\newblock Big bird: Transformers for longer sequences.
\newblock \emph{arXiv preprint arXiv:2007.14062}, 2020.

\bibitem[Zhai et~al.(2022)Zhai, Kolesnikov, Houlsby, and Beyer]{zhai2022scaling}
Xiaohua Zhai, Alexander Kolesnikov, Neil Houlsby, and Lucas Beyer.
\newblock Scaling vision transformers.
\newblock In \emph{Proceedings of the IEEE/CVF conference on computer vision and pattern recognition}, pages 12104--12113, 2022.

\bibitem[Zhang et~al.(2023{\natexlab{a}})Zhang, Han, Qiao, Kim, Bae, Lee, and Hong]{zhang2023faster}
Chaoning Zhang, Dongshen Han, Yu~Qiao, Jung~Uk Kim, Sung-Ho Bae, Seungkyu Lee, and Choong~Seon Hong.
\newblock Faster segment anything: Towards lightweight sam for mobile applications.
\newblock \emph{arXiv preprint arXiv:2306.14289}, 2023{\natexlab{a}}.

\bibitem[Zhang et~al.(2013)Zhang, Javed, and Shah]{zhang2013video}
Dong Zhang, Omar Javed, and Mubarak Shah.
\newblock Video object segmentation through spatially accurate and temporally dense extraction of primary object regions.
\newblock In \emph{Proceedings of the IEEE conference on computer vision and pattern recognition}, pages 628--635, 2013.

\bibitem[Zhang et~al.(2023{\natexlab{b}})Zhang, Cui, Wu, and Wang]{zhang2023joint}
Jiaming Zhang, Yutao Cui, Gangshan Wu, and Limin Wang.
\newblock Joint modeling of feature, correspondence, and a compressed memory for video object segmentation.
\newblock \emph{arXiv preprint arXiv:2308.13505}, 2023{\natexlab{b}}.

\bibitem[Zhang et~al.(2024{\natexlab{a}})Zhang, Wang, Gu, Li, Wang, Ling, and Tao]{zhang2024sam2}
Mingya Zhang, Liang Wang, Limei Gu, Zhao Li, Yaohui Wang, Tingshen Ling, and Xianping Tao.
\newblock Sam2-path: A better segment anything model for semantic segmentation in digital pathology.
\newblock \emph{arXiv preprint arXiv:2408.03651}, 2024{\natexlab{a}}.

\bibitem[Zhang et~al.(2023{\natexlab{c}})Zhang, Song, and Yao]{zhang2023deshadow}
Xiao~Feng Zhang, Tian~Yi Song, and Jia~Wei Yao.
\newblock Deshadow-anything: When segment anything model meets zero-shot shadow removal.
\newblock \emph{arXiv preprint arXiv:2309.11715}, 2023{\natexlab{c}}.

\bibitem[Zhang et~al.(2024{\natexlab{b}})Zhang, Cheng, Hu, Liu, Liu, Ran, Chen, Liu, and Wang]{zhang2024evf}
Yuxuan Zhang, Tianheng Cheng, Rui Hu, Lei Liu, Heng Liu, Longjin Ran, Xiaoxin Chen, Wenyu Liu, and Xinggang Wang.
\newblock Evf-sam: Early vision-language fusion for text-prompted segment anything model.
\newblock \emph{arXiv preprint arXiv:2406.20076}, 2024{\natexlab{b}}.

\bibitem[Zhao et~al.(2023)Zhao, Ding, An, Du, Yu, Li, Tang, and Wang]{zhao2023fast}
Xu~Zhao, Wenchao Ding, Yongqi An, Yinglong Du, Tao Yu, Min Li, Ming Tang, and Jinqiao Wang.
\newblock Fast segment anything.
\newblock \emph{arXiv preprint arXiv:2306.12156}, 2023.

\bibitem[Zhou et~al.(2024)Zhou, Sun, Li, Benini, and Konukoglu]{zhou2024sam2}
Yuli Zhou, Guolei Sun, Yawei Li, Luca Benini, and Ender Konukoglu.
\newblock When sam2 meets video camouflaged object segmentation: A comprehensive evaluation and adaptation.
\newblock \emph{arXiv preprint arXiv:2409.18653}, 2024.

\end{thebibliography}

\clearpage
\newpage
\beginappendix

% \clearpage
% \setcounter{page}{1}
% \maketitlesupplementary
\section{Efficient Cross-Attention}
Assume $\Tilde{K}_s$ is a coarser representation of memory spatial keys, $K_s$, a good surrogate of $K_s \in \R^{n \times d}$ with the same size, $\Bar{K}_s \in \R^{n \times d}$ from $\Tilde{K}_s \in \R^{\Tilde{w}\Tilde{h} \times d}$ is constructed by stacking each $\Tilde{k}_i, i = 1, \dots, \Tilde{w}\Tilde{h}$, $l_w\times l_h$ times, 
\begin{align*}
    \Bar{K}_s = [\underbrace{\Tilde{k}_1; \dots; \Tilde{k}_1}_{l_w\times l_h}; \underbrace{\Tilde{k}_2; \dots; \Tilde{k}_2}_{l_w\times l_h}; \dots; \underbrace{\Tilde{k}_{\Tilde{w}\Tilde{h}}; \dots; \Tilde{k}_{\Tilde{w}\Tilde{h}}}_{l_w\times l_h}]
\end{align*}
Each $\Tilde{v}_i, i =1, \dots, \Tilde{w}\Tilde{h}$, is stacked $l_w\times l_h$ times to make $\Bar{V}_s  \in \R^{n \times d}$ as a good surrogate of values, $V_s \in \R^{n\times d}$, 
\begin{align*}
    \small 
    \Bar{V}_s= [\underbrace{\Tilde{v}_1; \dots; \Tilde{v}_1}_{l_w\times l_h}; \underbrace{\Tilde{v}_2; \dots; \Tilde{v}_2}_{l_w \times l_h}; \dots; \underbrace{\Tilde{v}_{\Tilde{w}\Tilde{h}}; \dots; \Tilde{v}_{\Tilde{w}\Tilde{h}}}_{l_w \times l_h}]
\end{align*}
 The concatenation of coarse spatial tokens with object pointer tokens is, $\Bar{K} = [\Bar{K}_s; K_p]\in \R^{(n+P)\times d}$ and $\Bar{V} = [\Bar{V}_s; V_p]\in \R^{(n+P)\times d}$. 
 \begin{lemma}\label{lem:equiv}
For the coarse memory tokens, $\bar{K}$ and $\bar{V}$,  queries $Q\in\R^{L\times d}$, we have,
\begin{align}\label{eq:replace}
    \text{softmax}\Lleft\frac{Q\Bar{K}^{T}}{\sqrt{d}}\Rright\Bar{V} = \text{softmax}\Lleft A \Rright\Tilde{V},
\end{align}
where $A = [\frac{Q\Tilde{K}_s^{T}}{\sqrt{d}} + \ln{(l_w\times l_h)}, \frac{QK_p^{T}}{\sqrt{d}}] \in \R^{L\times (\Tilde{w}\Tilde{h} + P)}$, $\Tilde{V} = [\Tilde{V}_s; V_p] \in \R^{(\Tilde{w}\Tilde{h}+P)\times d}$.
\end{lemma}
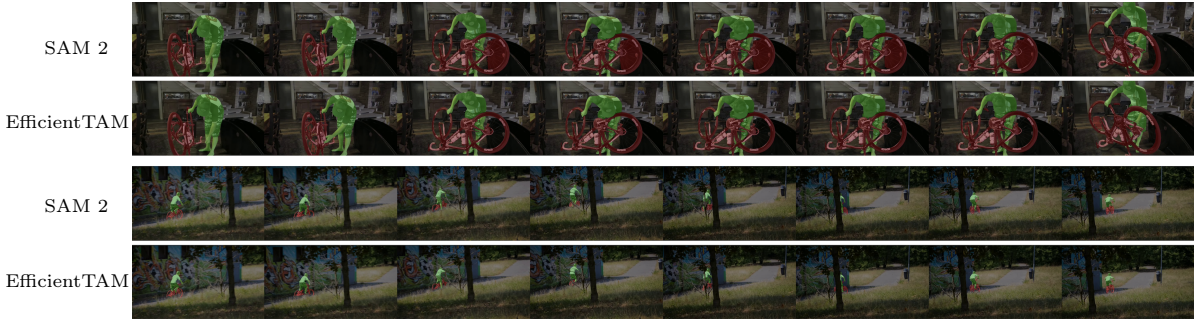
\begin{figure*}[t!]
\centering
% \vspace{5pt}
\begin{overpic}[width=0.85\linewidth]{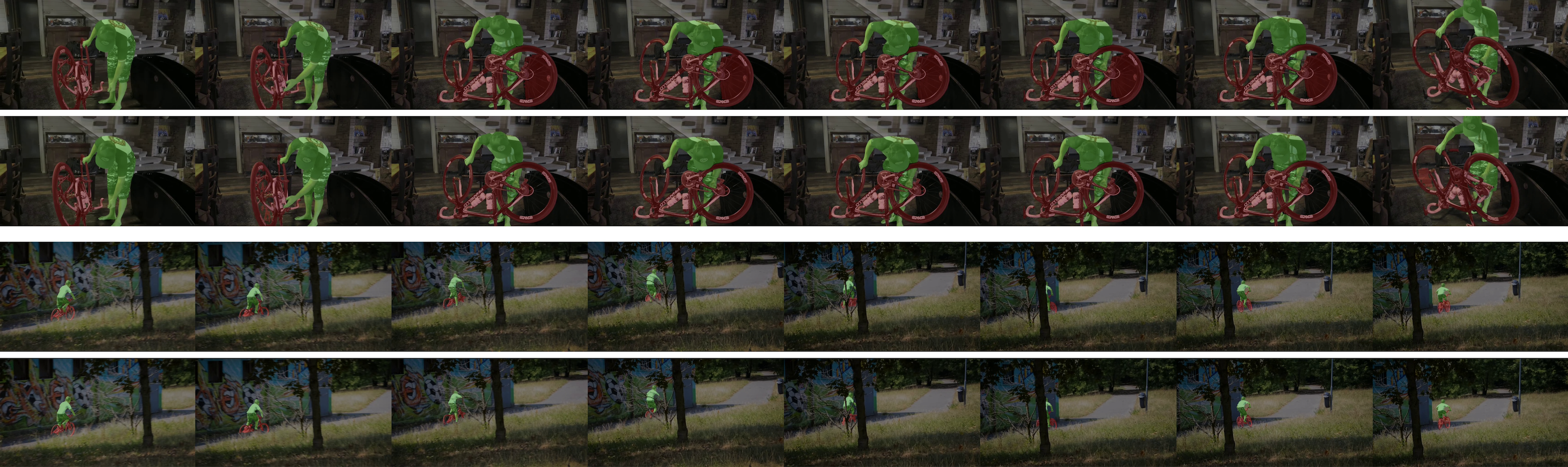}
\put (-8.3,25) {\scriptsize{SAM 2}}
\put (-12,18) {\scriptsize{EfficientTAM}}
\put (-8.3,10) {\scriptsize{SAM 2}}
\put (-12,3) {\scriptsize{EfficientTAM}}
\end{overpic}
\caption{Visualization results on video segmentation and tracking with SAM 2, and our EfficientTAM model. We sampled a subset of frames for visualization. The segmented objects with occlusion are colored in green. }
\label{fig:visual_vost3}
\end{figure*}
\begin{proof}
Denote $Q = [q_1; \dots; q_L]$, where $q_i \in \R^{1 \times d}$. The cross-attention matrix, $\Bar{C} = \text{softmax}\Lleft\frac{Q\Bar{K}^{T}}{\sqrt{d}}\Rright\Bar{V} \in \R^{L\times d}$. The softmax matrix $\bar{S} = \text{softmax}\Lleft\frac{Q\Bar{K}^{T}}{\sqrt{d}}\Rright \in \R^{L\times (n+P)}$ can be formulated as, 
\begin{equation*}
\resizebox{0.7\linewidth}{!}{$
\bar{S} = D_{\mathcal{S}}\begin{bmatrix} 
    e(\frac{q_1}{\sqrt{d}}\Tilde{k}_1^T) & \dots & e(\frac{q_1}{\sqrt{d}}\Tilde{k}_1^T) &\dots & e(\frac{q_1}{\sqrt{d}}\Tilde{k}_{\Tilde{w}\Tilde{h}}^T) & \dots & e(\frac{q_1}{\sqrt{d}}K_p^T)\\
    \vdots & \dots & \vdots & \dots & \vdots &\dots & \dots \\
    e(\frac{q_L}{\sqrt{d}}\Tilde{k}_1^T) & \dots  & e(\frac{q_L}{\sqrt{d}}\Tilde{k}_1^T) &\dots & e(\frac{q_L}{\sqrt{d}}\Tilde{k}_{\Tilde{w}\Tilde{h}}^T) & \dots & e(\frac{q_L}{\sqrt{d}}K_p^T)
    \end{bmatrix}$}
\end{equation*}
where $D_{\mathcal{S}}$ is a $L\times L$ diagonal matrix, which normalizes each row of the $\bar{S}$ matrix such that the row entries sum up to 1, and $e(\cdot)$ denotes $\exp(\cdot)$. 
For each row of the cross-attention matrix, we have, 
\begin{align}\label{eq:entry}
    \Bar{C}_{ij} & = D_{{\mathcal{S}}_{ii}}(\underbrace{e(\frac{q_i}{\sqrt{d}}\Tilde{k}_1^T)\Tilde{v}_1 + \dots e(\frac{q_i}{\sqrt{d}}\Tilde{k}_1^T)\Tilde{v}_1}_{l_w\times l_h} 
     + \dots + \underbrace{e(\frac{q_i}{\sqrt{d}}\Tilde{k}_1^T)\Tilde{v}_{\Tilde{w}\Tilde{h}}  + \dots e(\frac{q_i}{\sqrt{d}}\Tilde{k}_1^T)\Tilde{v}_{\Tilde{w}\Tilde{h}}}_{l_w\times l_h}  + e(\frac{q_i}{\sqrt{d}}K_p^T)V_p) \nonumber \\ 
    &= D_{{\mathcal{S}}_{ii}}(l_w\times l_h\times (e(\frac{q_i}{\sqrt{d}}\Tilde{k}_1^T)\Tilde{v}_1 + \dots +  e(\frac{q_i}{\sqrt{d}}\Tilde{k}_1^T)\Tilde{v}_{\Tilde{w}\Tilde{h}}) + e(\frac{q_i}{\sqrt{d}}K_p^T)V_p) \nonumber \\ 
    & = D_{{\mathcal{S}}_{ii}}(l_w\times l_h \times e(\frac{q_i}{\sqrt{d}}\Tilde{K}_s^T)\Tilde{V}_s^T + e(\frac{q_i}{\sqrt{d}}K_p^T)V_p) \nonumber \\ 
    & = D_{{\mathcal{S}}_{ii}}(e(\ln(l_w\times l_h) + \frac{q_i}{\sqrt{d}}\Tilde{K}_s^T)\Tilde{V}_s + e(\frac{q_i}{\sqrt{d}}K_p^T)V_p) \nonumber \\ 
    & = \text{softmax}[\frac{q_i\Tilde{K}_s^{T}}{\sqrt{d}} + \ln{(l_w\times l_h)}, \frac{q_i\Tilde{K}_p^{T}}{\sqrt{d}}][\Tilde{V}_s; V_p] 
\end{align}
where $D_{{\mathcal{S}}_{ii}}$ is the $i^{\text{th}}$ diagonal element of the matrix $D_{\mathcal{S}}$. Note that the right side of \cref{eq:entry} is the $i^{\text{th}}$ row of $\text{softmax}\Lleft A \Rright\Tilde{V}$.
It concludes the proof. 
\end{proof}

% \input{tables/vos}
% \section{Main Results}
% We notice one typo of Tab. 1 in the main paper. The LVOS val number of SAM 2 did not show up in the table. We fixed this typo in \cref{tab:vos}.
\section{Ablation Studies}

\begin{table}[t]
    \centering
    \resizebox{0.65\linewidth}{!}{
    \begin{tabular}{c|ccc}
    \hline
    Object Pointers & MOSE dev & DAVIS 2017 val & SA-V test \\ \hline
    No             & 75.8     & 89.0             & 72.1      \\ \hline
    \rowcolor{gray} Yes              & 76.5     & 89.2           & 74.5      \\ \hline
    \end{tabular}}
    \caption{\centering Ablation study on the design of memory cross-attention in EfficientTAM.}
    \label{tab:pointer}
\end{table}
\textbf{Impact of the object pointer tokens.} We study the effect of the object pointer tokens when performing cross-attention in the memory module. We ablate the cross-attention with or without the object pointer tokens. When performing cross-attention,  
we find that object pointers significantly improve the performance on SA-V test dataset, 74.5 vs 72.1 $\mathcal{J}$\&$\mathcal{F}$, shown in \cref{tab:pointer}. The observations are consistent with SAM 2~\citep{ravi2024sam}. 
This demonstrates that object pointer tokens need to be cross-attended with spatial tokens.

\begin{table}[t]
    \centering
    \resizebox{0.65\linewidth}{!}{
    \begin{tabular}{c|ccc}
    \hline
    Pooling        & MOSE dev & DAVIS 2017 val & SA-V test \\ \hline
    Memory tokens & 74.5     & 87.6           & 71.7      \\ \hline
    \rowcolor{gray} Spatial tokens only  & 76.5     & 88.6           & 74.0        \\ \hline
    \end{tabular}}
    \caption{Ablation study on taking care of the memory token structure for efficient cross-attention in EfficientTAM.}
    \label{tab:pooling}
\end{table}
\noindent \textbf{Structure of memory tokens.} We ablate the impact of memory tokens for efficient cross-attention in the memory module. In our efficient cross-attention, we leverage the locality of memory spatial tokens for a coarser representation, and we concatenate the coarser embedding with object pointer tokens. 
In \cref{tab:pooling}, we observe that naively pooling the entire memory tokens instead of only the spatial tokens yields a large performance drop, 2.3 $\mathcal{J}$\&$\mathcal{F}$ on SA-V test.

\begin{table}[t]
    \centering
    \resizebox{0.65\linewidth}{!}{
    \begin{tabular}{c|ccc}
    \hline
    Cross-Attention & MOSE dev & DAVIS 2017 val & SA-V test \\ \hline
    Local-windowed  & 75.4     & 88.6           & 72.4      \\ \hline
    \rowcolor{gray} Pooling   & 76.5     & 88.6           & 74.0        \\ \hline
    \end{tabular}}
    \caption{\centering Comparing  with local windowed attention.}
    \label{tab:windowed}
\end{table}
\noindent \textbf{Local windowed cross-attention.} We adapt local windowed attention for efficient cross-attention by partitioning input tokens into 4 non-overlapping segments (windows), within which we conduct cross-attention. In \cref{tab:windowed}, we find that local windowed cross-attention underperforms our proposed efficient cross-attention using averaging pooling, 72.4 vs 74.0 $\mathcal{J}$\&$\mathcal{F}$ on SA-V test dataset. These results demonstrate the effectiveness of our efficient cross-attention by leveraging the strong locality of spatial memory tokens. 

\begin{figure}[t]
    \centering
    \begin{overpic}[width=0.5\linewidth]{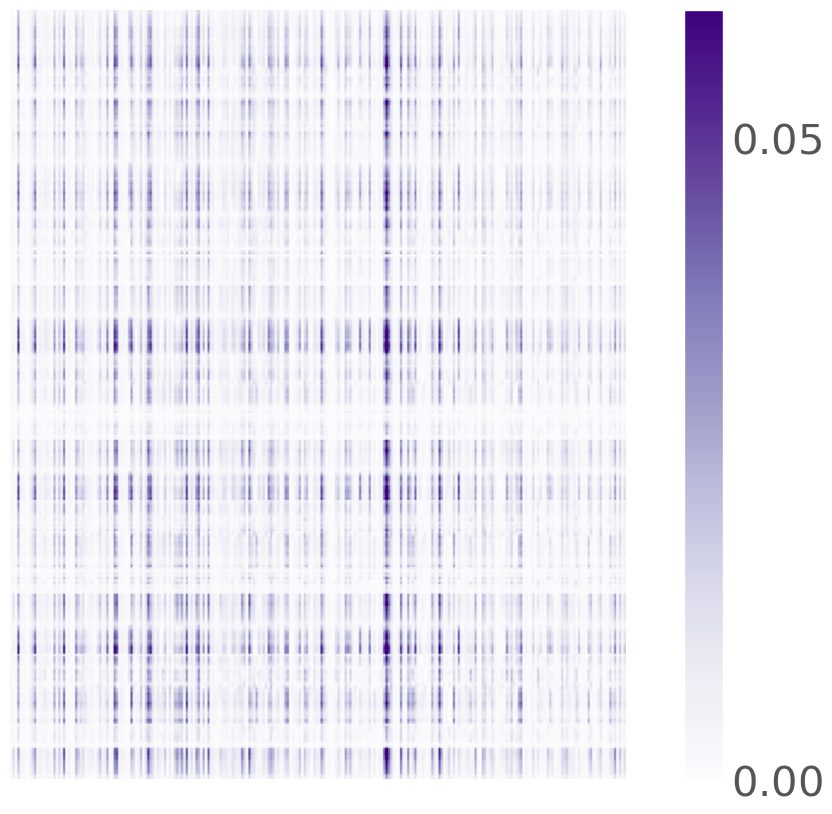}
\end{overpic}
   \vspace{-5pt}
    \caption{Visualization of the difference between original cross-attention and efficient cross-attention of \cref{eq:acrossattn}.}
    \label{fig:across_attn}
\end{figure}

\noindent \textbf{Efficient cross-attention variant.} We observe that \cref{eq:acrossattn} in the main paper is close to original cross-attention, visualized in \cref{fig:across_attn}. This suggests that \cref{eq:acrossattn} can also serve as a surrogate of the original cross-attention. 

\section{Qualitative Evaluation}
\begin{figure}
\vspace{10pt}
    \centering
    \begin{overpic}[width=1.0\linewidth]{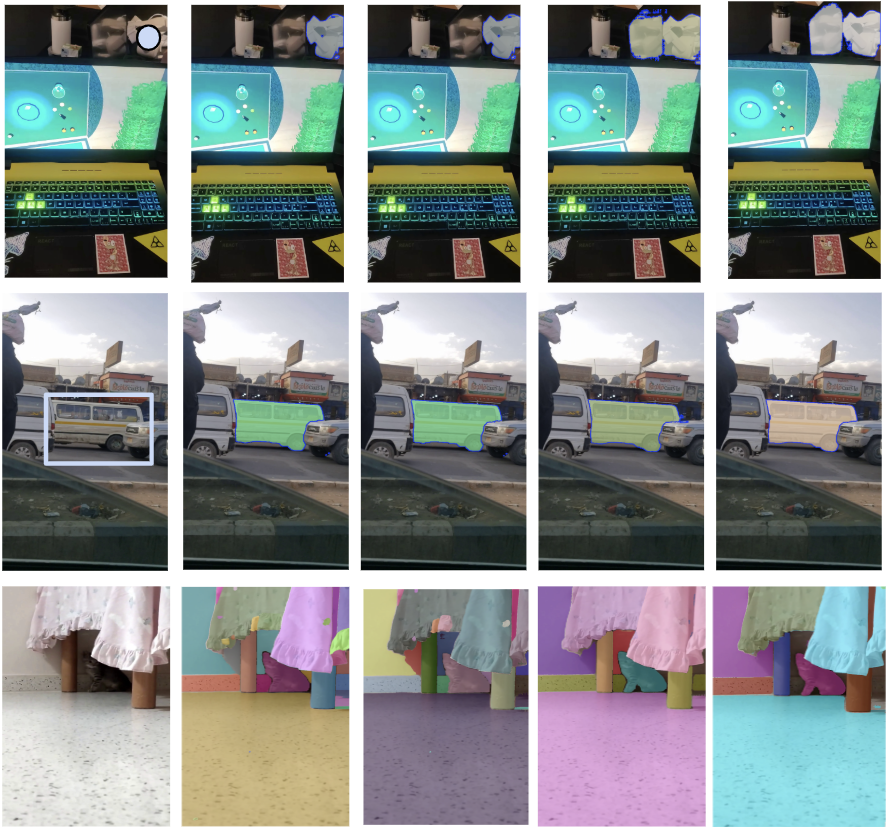}
\put (5,96) {\scriptsize{Input Image}}
\put (17,96) {\scriptsize{SAM\citep{kirillov2023segment}}}
\put (37,96) {\scriptsize{EfficientSAM\citep{xiong2024efficientsam}}}
\put (63,96) {\scriptsize{SAM 2\citep{ravi2024sam}}}
\put (85,96) {\scriptsize{EfficientTAM}}
\end{overpic}
    \caption{Visualization results on image segmentation with point-prompt, box-prompt, and segment everything for SAM, EfficientSAM, SAM 2, and our EfficientTAM model.}
    \label{fig:visual_seg}
\end{figure}
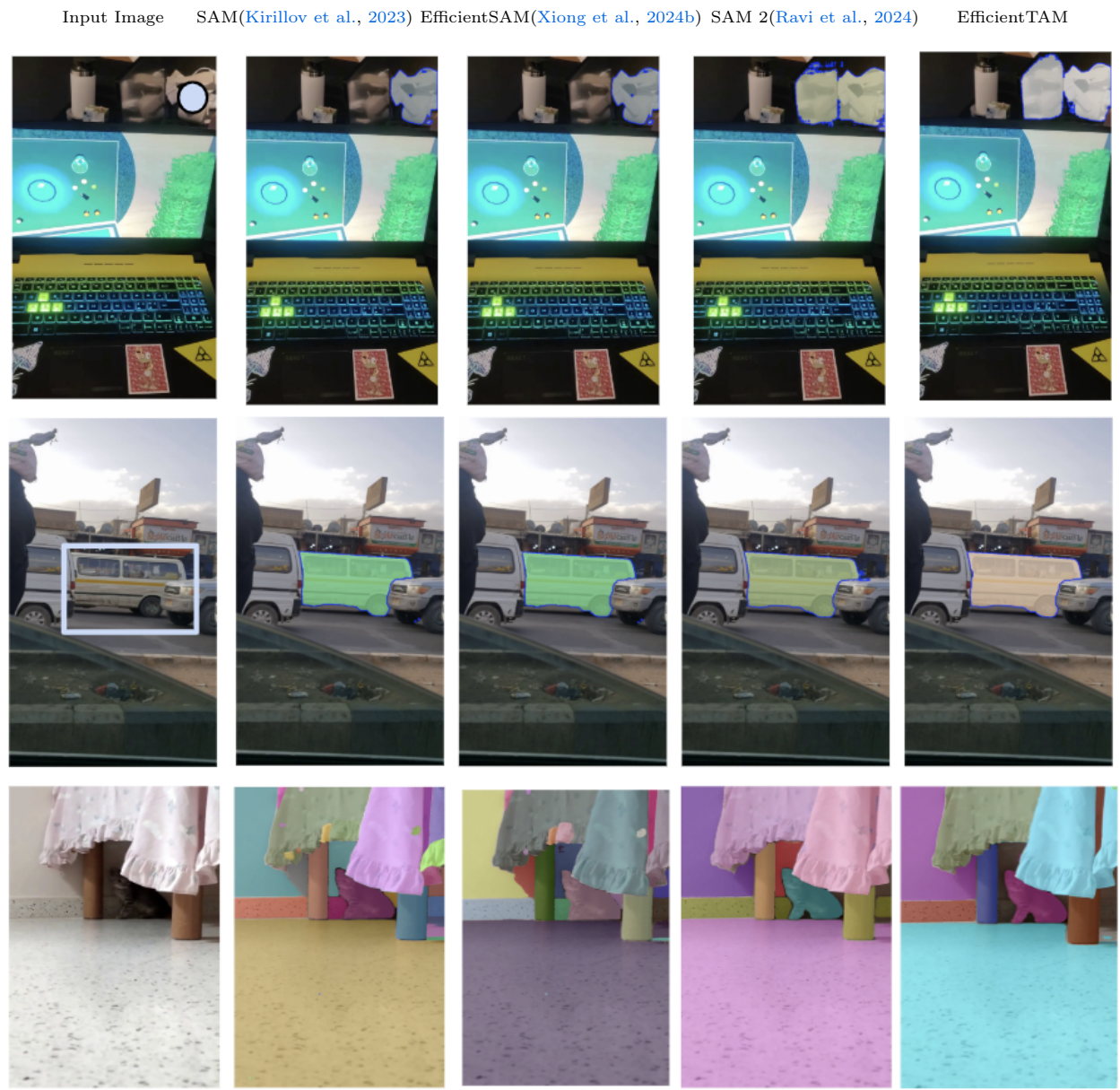

We provide more qualitative results of EfficientTAMs for video and image instance segmentation. \cref{fig:visual_vost3} shows two challenging video examples with occluded objects. We compare EfficientTAM and SAM 2 with a mask in the first frame prompted. We find that our EfficientTAM can generate high-quality masklet for the target occluded object as SAM 2. For image segmentation, we also observe that our EfficientTAM can generate quality image segmentation results as SAM and SAM 2, shown in  \cref{fig:visual_seg}. We report the predicted masks with two types of prompts, point and box, and also segment everything results. These results suggest that our EfficientTAMs have similar abilities to SAM 2, while EfficientTAM is more efficient. 

\end{document}